\documentclass[10pt,journal,compsoc]{IEEEtran}
\usepackage{amsmath,amsfonts}
\usepackage{algorithmic}
\usepackage{algorithm}
\usepackage{array}
\usepackage[caption=false,font=normalsize,labelfont=sf,textfont=sf]{subfig}
\usepackage{textcomp}
\usepackage{stfloats}
\usepackage{url}
\usepackage{verbatim}
\usepackage{graphicx}
\usepackage{cite}
\usepackage{xspace}
\usepackage{threeparttable}
\usepackage{xcolor}
\usepackage{multirow}
\usepackage[T1]{fontenc}
\usepackage{gensymb}
\usepackage[T1]{fontenc}
\usepackage{booktabs}
\usepackage{amssymb}
\usepackage{enumitem}
\usepackage{hyperref}
\usepackage[hyphenbreaks]{breakurl}
\usepackage{mathtools}
\usepackage{xspace}
\usepackage{diagbox}
\usepackage{graphbox}

\usepackage[capitalize]{cleveref}
\crefname{section}{Sec.}{Secs.}
\Crefname{section}{Section}{Sections}
\Crefname{table}{Table}{Tables}
\crefname{table}{Tab.}{Tabs.}

\hypersetup{hidelinks} 

\makeatletter
\newcommand*{\rom}[1]{\expandafter\@slowromancap\romannumeral #1@}
\makeatother

\newcommand{\ie}{{\emph{i.e.}}\xspace}
\newcommand{\eg}{{\emph{e.g.}}\xspace}

\newcommand{\etal}{{\emph{et al.}}}
\newcommand{\FLIP}{\protect\reflectbox{F}LIP\xspace}

\begin{document}

\title{Measuring Perceptual Color Differences of Smartphone Photographs}

\author{Zhihua~Wang, 
Keshuo~Xu, 
Yang~Yang,
Jianlei~Dong,
Shuhang~Gu,
Lihao~Xu, 
Yuming~Fang,~\IEEEmembership{Senior Member,~IEEE}, and
Kede~Ma,~\IEEEmembership{Senior Member,~IEEE}
\IEEEcompsocitemizethanks{
\IEEEcompsocthanksitem Z. Wang is with the Guangdong Laboratory of Machine Perception and Intelligent Computing, Shenzhen MSU-BIT University, China, and also with the Department of Computer Science, City University of Hong Kong, Kowloon, Hong Kong (e-mail: zhihua.wang@my.cityu.edu.hk). 
\IEEEcompsocthanksitem K. Xu and Y. Fang are with the School of Information Management, Jiangxi University of Finance and Economics, Nanchang, China (e-mail: gliterxu@gmail.com, fa0001ng@e.ntu.edu.sg). 
\IEEEcompsocthanksitem Y. Yang and J. Dong are with the Guangdong OPPO Mobile Telecommunications Corp., Ltd.  (e-mail: yangyang5@oppo.com, dongjianlei@oppo.com). 
\IEEEcompsocthanksitem S. Gu is with the School of Computer Science and Engineering,
University of Electronic Science and Technology of China (e-mail: shuhanggu@gmail.com).
\IEEEcompsocthanksitem L. Xu is with the School of Digital Media and Art Design,
Hangzhou Dianzi University, Hangzhou, China (e-mail: ssyjh123@hdu.edu.cn). 
\IEEEcompsocthanksitem K. Ma is with the Department of Computer Science, City University of Hong Kong, Kowloon, Hong Kong (e-mail: kede.ma@cityu.edu.hk). 
}
\thanks{Corresponding author: Kede Ma.}
}

\markboth{IEEE Transactions on Pattern Analysis and Machine Intelligence}%
{Shell \MakeLowercase{\textit{et al.}}: A Sample Article Using IEEEtran.cls for IEEE Journals}
\IEEEtitleabstractindextext{%
\begin{abstract}
Measuring perceptual color differences (CDs) is of great importance in modern smartphone photography. Despite the long history, most CD measures have been constrained by psychophysical data of homogeneous color patches or a limited number of simplistic natural photographic images. It is thus questionable whether existing CD measures generalize in the age of smartphone photography characterized by greater content complexities and learning-based image signal processors. 
In this paper, we put together so far the largest image dataset for perceptual CD assessment, in which the photographic images are 1) captured by six flagship smartphones, 2) altered by Photoshop\textsuperscript{\textregistered}, 3) post-processed by built-in filters of the smartphones, and 4) reproduced with incorrect color profiles. We then conduct a large-scale psychophysical experiment to gather perceptual CDs of $30,000$ image pairs in a carefully controlled laboratory environment. Based on the newly established dataset, we make one of the first attempts to construct an end-to-end learnable CD formula based on a lightweight neural network, as a generalization of several previous metrics. Extensive experiments demonstrate that the optimized formula outperforms $33$ existing CD measures by a large margin, offers 
reasonable local CD maps without the use of dense supervision, generalizes well to homogeneous color patch data, and empirically behaves as a proper metric in the mathematical sense. Our dataset and code are publicly available at \url{https://github.com/hellooks/CDNet}.
\end{abstract}

}

\maketitle

\IEEEdisplaynontitleabstractindextext

\IEEEpeerreviewmaketitle

\begin{figure}[t]
    \vspace{-1.2cm}
    \begin{minipage}[t]{0.52\textwidth}
        \hskip-1.6em
        \centering
        \subfloat[]{\includegraphics[width=0.32\textwidth]{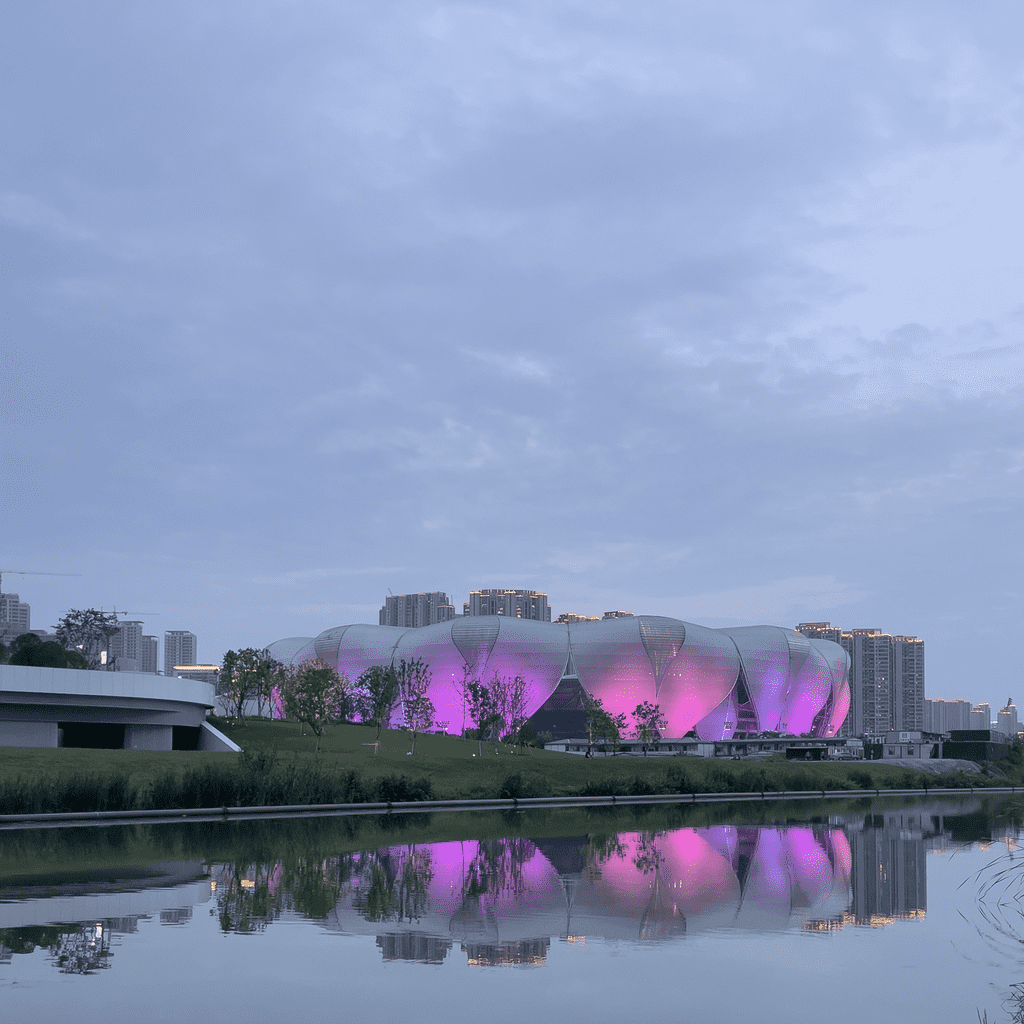}}\hskip.4em
        \subfloat[]{\includegraphics[width=0.32\textwidth]{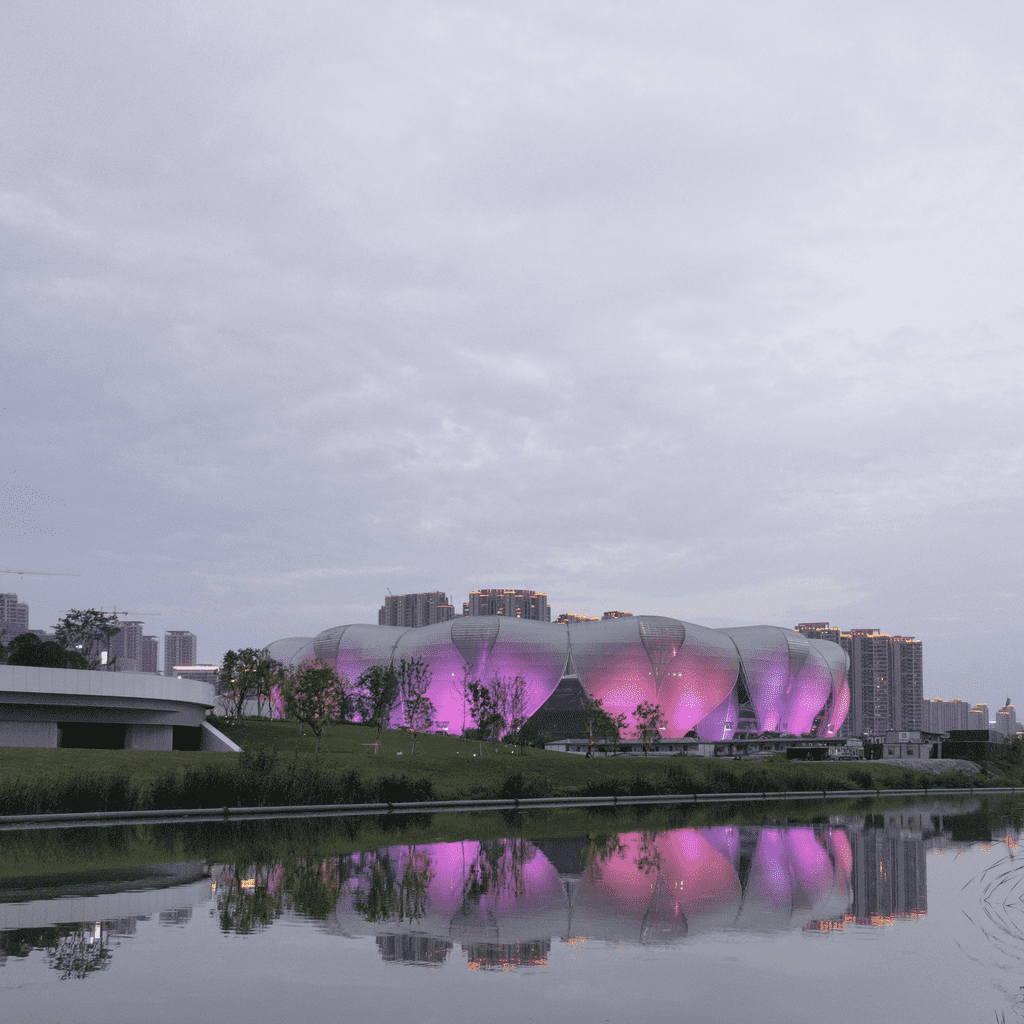}}\hskip.4em
        \subfloat[]{\includegraphics[width=0.32\textwidth]{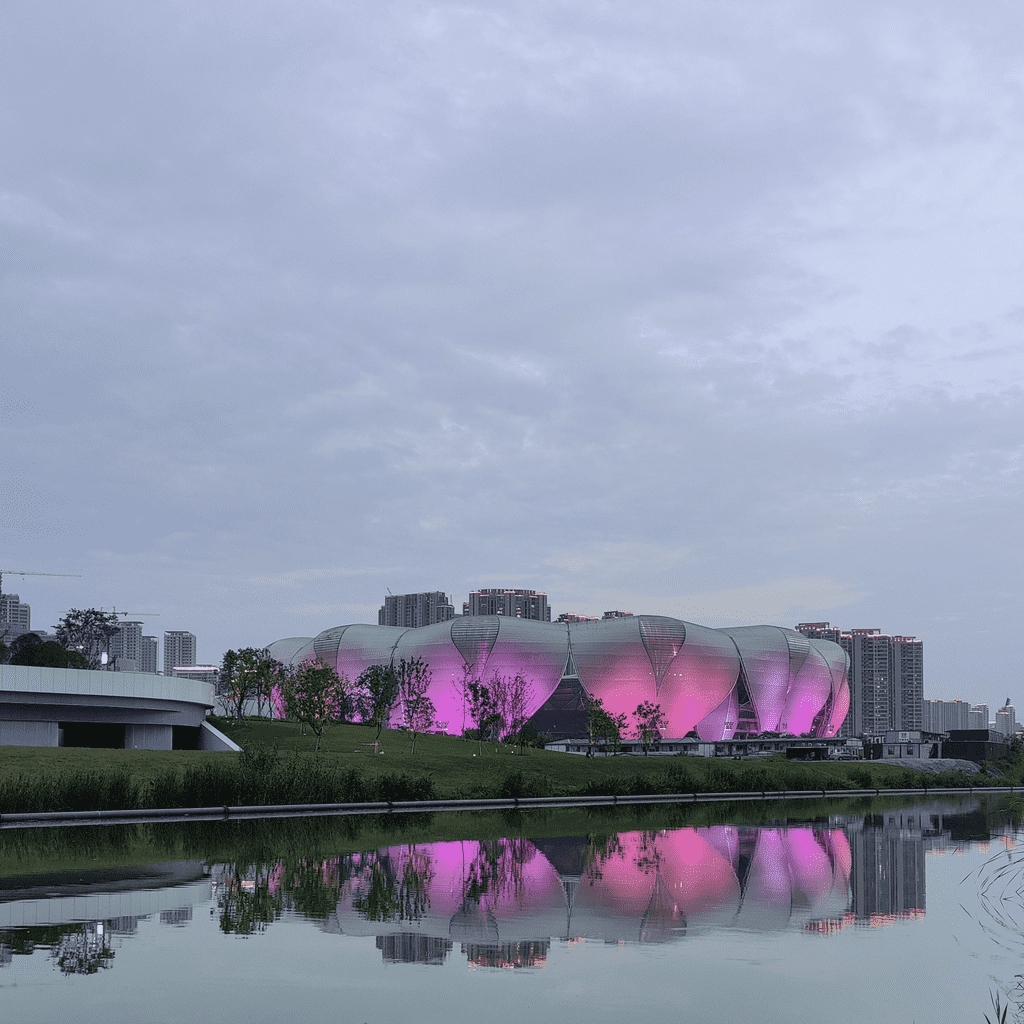}}
        \vspace{-0.2cm}
    \end{minipage}

    \begin{minipage}[t]{0.52\textwidth}
        \hskip-1.6em
        \centering
        \subfloat[]{\includegraphics[width=0.32\textwidth]{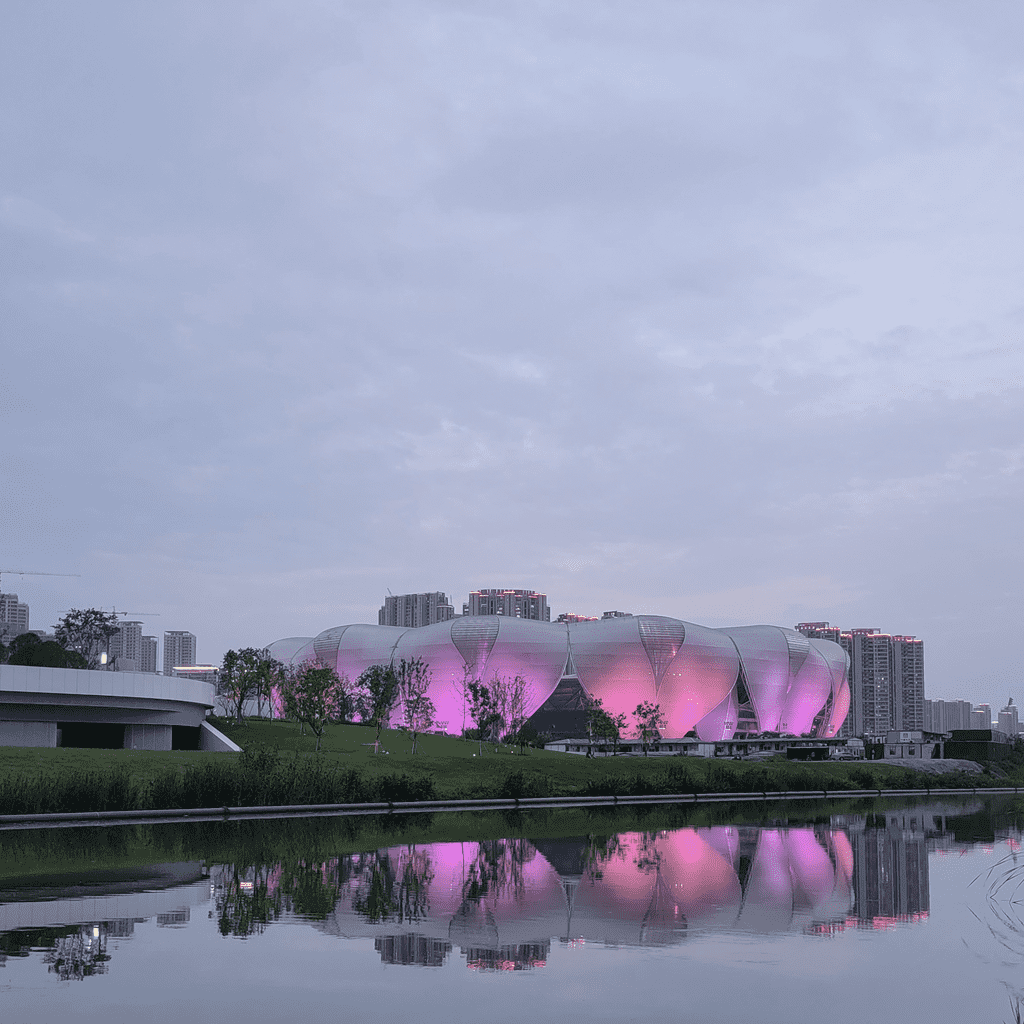}}\hskip.4em 
        \subfloat[]{\includegraphics[width=0.32\textwidth]{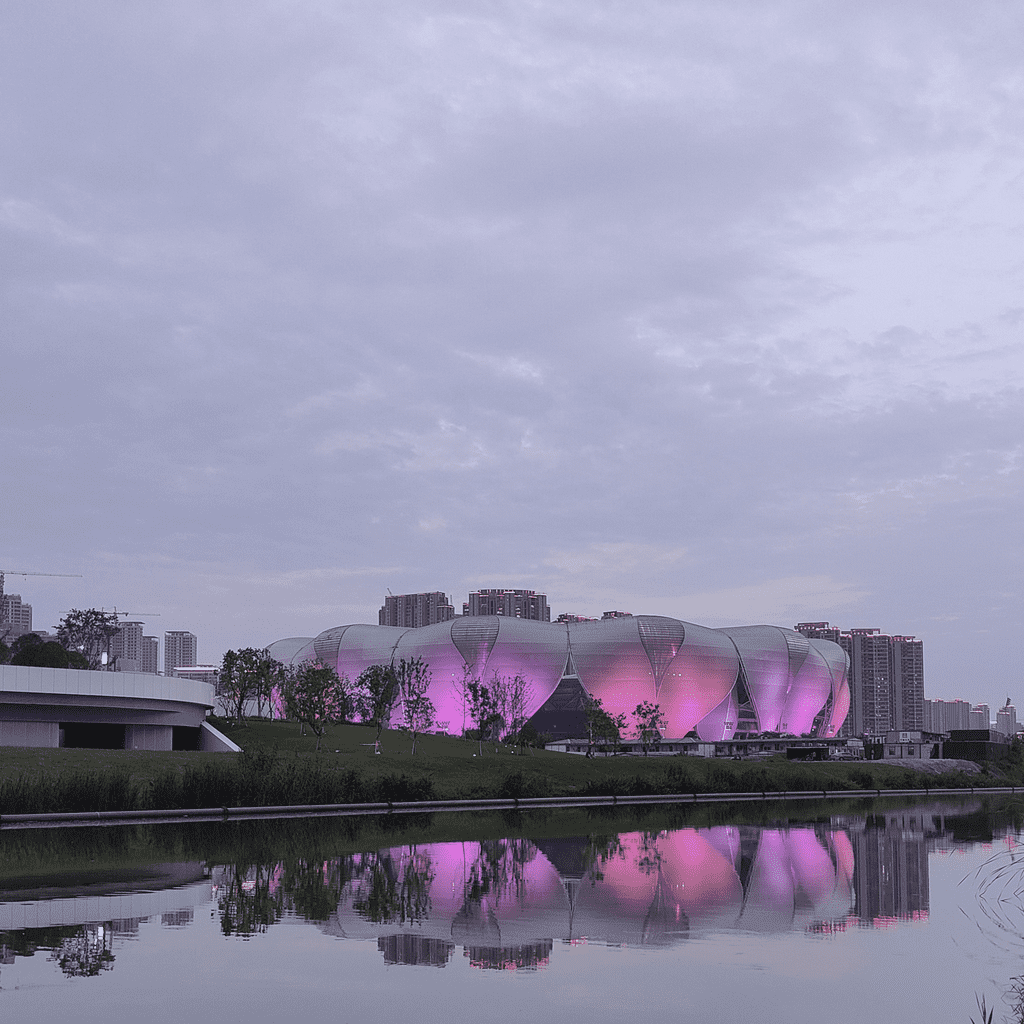}}\hskip.4em 
        \subfloat[]{\includegraphics[width=0.32\textwidth]{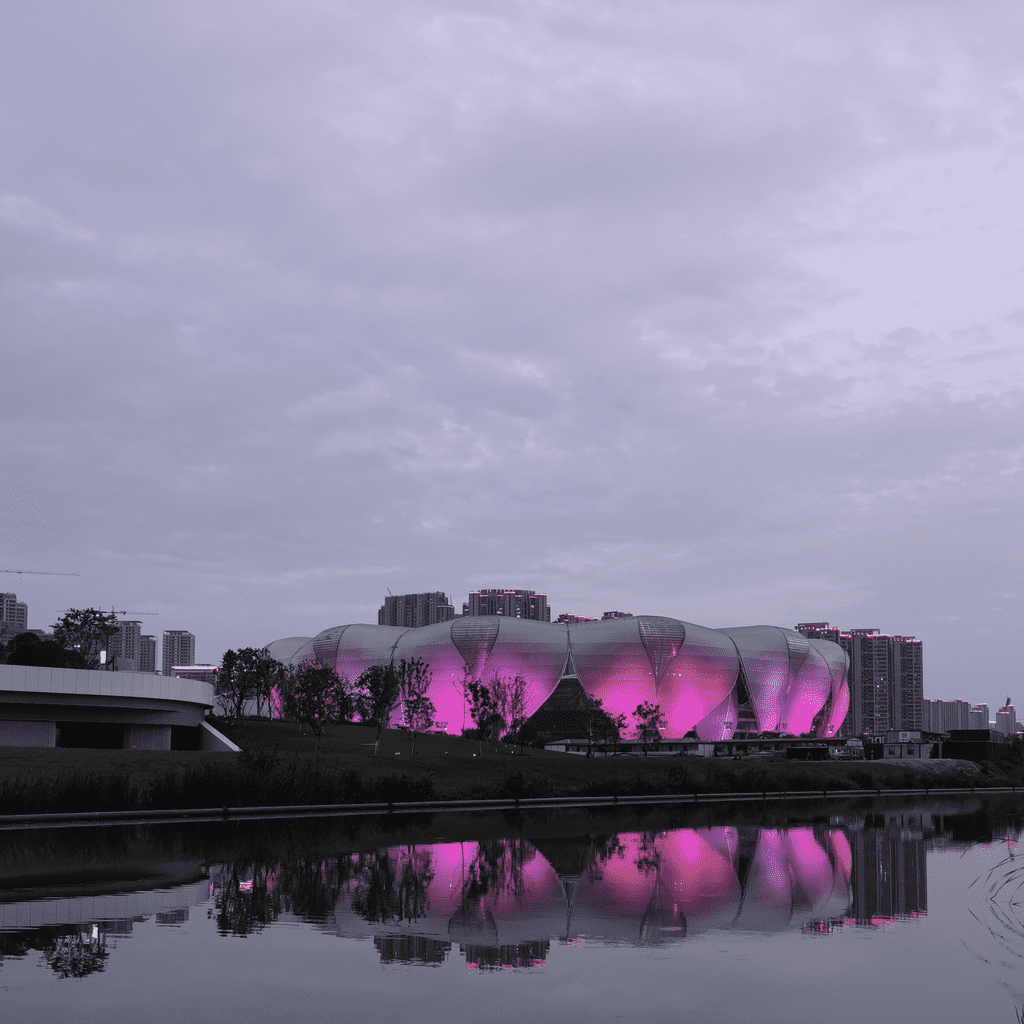}}
    \end{minipage}
    \caption{Sample pictures of the same natural scene captured by six flagship smartphones using the night mode. It is clear that they reproduce similar structural details but different color appearances. (a) Apple iPhone 12 Pro. (b) HUAWEI Mate40 Pro. (c) OnePlus 7 Pro. (d) Samsung S21 Ultra. (e) OPPO Find X3 Pro. (f) Xiaomi 11 Ultra.}
    \label{fig:color_difference}
\end{figure}

\IEEEraisesectionheading{\section{Introduction}
\label{sec:intro}}
\IEEEPARstart{N}{owadays}, a smartphone is more of a camera than a phone~\cite{delbracia2021mobile}. Due to its convenience and flexibility, smartphones have become the standard digital device for most casual photography. Arguably the primary selling point of a smartphone is its picture-taking quality, which spurs the manufacturers to upgrade the camera system and the associated image signal processor (ISP) at an accelerated pace. Picture quality is determined by a weighted combination of all visually significant attributes~\cite{wang2006modern}, among which color plays an increasingly important role. This is because the flagship smartphones on the market are able to reproduce the structural details for the majority of natural scenes; it is the color appearance that differentiates them (see Fig.~\ref{fig:color_difference}). Even in the case of poor lighting conditions where the scene details are not lit properly, current ISPs are able to do magical wonders to fill in plausible things and stuff~\cite{adelson2001seeing} in the scene. Again, the resulting images may then be different mainly in color reproduction.  

\begin{figure*}[t]
    \includegraphics[width=1.00\textwidth]{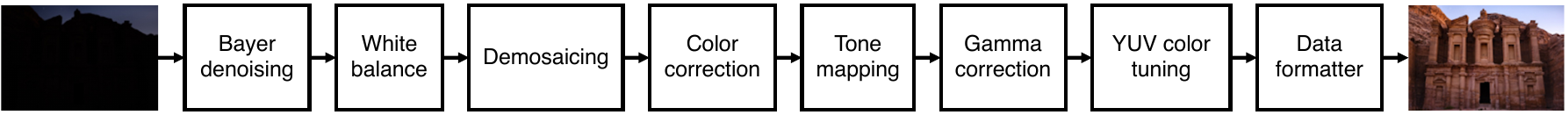}
    \caption{Major components of the ISP pipeline.}
    \label{fig:isp_pipeline}
\end{figure*}
Color is not merely a physical property associated with an object. It is a visual sensation that may be affected by luminance, viewing conditions, and the state of the eye's adaptation~\cite{edition2004colorimetry}. The history of color perception and production has been written by both the greatest scientists and artists, including Newton who discovered the visible spectrum of light and created the first color wheel, Goethe who marked the beginning of studying the perception and psychological impact of color, Young and Helmholtz whose trichromatic theory foreshadowed the modern understanding of color vision, and Munsell who defined color in terms of value, chroma, and hue.  

In modern smartphone photography, the most important aspect of color is \textit{color quality} (\eg, \textit{color preference}), which is, however, highly subjective and culturally conditioned. Thus, the difference or distance between two colors is a metric of interest in colorimetry. A common theme in designing color difference (CD) formulae is the pursuit of the perceptually uniform color space, in which the Euclidean distance between two points provides an accurate approximation of perceived CD. In 1976, the Commission Internationale de l’{\`E}clairage
(International Commission on Illumination, CIE) suggested two ``uniform'' color spaces, CIELAB and CIELUV for reflective and self-luminous colors, respectively~\cite{edition2004colorimetry}. It was soon discovered that these two are not uniform in several aspects, motivating non-Euclidean remedies, such as CMC~\cite{bsi1998cmc}, CIE94~\cite{mcdonald1995cie94}, and CIEDE2000~\cite{luo2001ciede2000}. However, these CD formulae were derived and tested on homogeneous color patches (of small CDs) against gray backgrounds\footnote{A typical viewing field consists of a stimulus and background that subtend visual angles} of $4{\degree}$ and $10{\degree}$, respectively~\cite{fairchild2013color}., which may not generalize to natural photographic images of complex content and color variations. Zhang and Wandell~\cite{zhang1997scielab} described a spatial extension to CIELAB CD metrics by incorporating a low-pass filter that simulates the point spread function of the eye optics. Recent studies~\cite{jaramillo2019evaluation} showed that the CD sensation of natural photographic images is more relevant to comparing color appearances of homogeneous textured surfaces. Nevertheless, most CD measures are evaluated using a very limited number of hand-picked examples altered by linear and quasi-linear color transforms~\cite{jaramillo2019evaluation}. Thus, their generalization to smartphone photographs in the real world is unclear, where CDs may arise from 
the differences in hardware imaging systems, ISPs, post-filters, and display configurations. 

To measure the progress of CD assessment and to facilitate the development of generalizable CD metrics, we carry out so far the most comprehensive CD study for smartphone photography. Our main contributions are fourfold.

\begin{itemize}
\item We construct the largest image dataset for color differences, which we name Smartphone Photography Color Difference (SPCD) dataset. It contains $15,335$ photographic images 1) captured by six flagship smartphones\footnote{The selected smartphones were the latest generations of their respective brands at the start date of this project.}, 2) altered by Photoshop\textsuperscript{\textregistered} to simulate ISP functions, 3) post-processed by built-in filters of one of the six smartphones, and 4) reproduced with incorrect International Color Consortium (ICC) color profiles. The images are chosen to span a variety of real-world picture-taking scenarios in terms of foreground diversity, background complexity, lighting and weather conditions, and camera mode.
  \item We carry out a large-scale psychophysical experiment to collect human judgments of perceptual CDs. We assemble $30,000$ image pairs, each of which is displayed on two color-calibrated monitors and rated by a panel of at least $20$ subjects in a well-controlled laboratory environment. After outlier detection and subject rejection, more than $600,000$ valid human ratings are received.
  \item We conduct a comprehensive performance comparison of $33$ existing CD measures on the proposed SPCD dataset, elucidating their relative advantages and disadvantages in assessing perceptual CDs. 
  \item We take initial steps to develop an end-to-end learnable CD method based on a lightweight deep neural network (DNN), that generalizes several existing CD metrics built on CIE colorimetry. Extensive experiments show that our method significantly outperforms existing CD measures, offers competitive local CD maps without dense supervision, exhibits reasonable generalization to homogeneous color patch data~\cite{luo2001ciede2000}, and empirically behaves as a proper metric mathematically.
  
\end{itemize}

\section{Related Work}
In this section, we review previous work that is closely related to ours, including color image pipelines, CD measures, and human-rated datasets for CD assessment.
\subsection{Color Image Pipelines} 
Color image pipelines implemented in ISPs are critical components for smartphone photography to convert RAW sensor data to JPEG images~\cite{kim2012new}. Knowledge-driven ISPs involve a sequence of modular operations to perform specific tasks. Fig.~\ref{fig:isp_pipeline} illustrates the major components of a typical knowledge-driven ISP, including 1) Bayer denoising, removing sensor noise from raw Bayer color patterns; 2) auto white balance, adjusting a picture to neutral (\ie, to make the white look white); 3) demosaicing, interpolating missing
color values from nearby pixels; 4) color correction, reducing color errors due to differences between the spectral characteristics of the image sensor and the spectral responses of the human eye; 5) tone mapping, adapting to the high-dynamic-range (HDR) imaging scenarios; 6) gamma correction, encoding and decoding the light and color information closer to what humans perceive; and 7) optional YUV color tuning, correcting color errors in YUV color space. In recent years, the optimization of data-driven ISPs becomes an active research area, and there is a surge of interest to replace the mobile camera ISP with DNNs~\cite{ignatov2020replacing}. 

Generally, the performance of ISPs is evaluated by general-purpose image quality models such as the peak signal-to-noise ratio (PSNR) and the structural similarity (SSIM) index~\cite{wang2004image}, and conventional CD metrics S-CIELAB~\cite{zhang1997scielab} and CIEDE2000~\cite{luo2001ciede2000}. Unfortunately, these measures fail to provide accurate and reliable CD assessment of natural photographic images, and thus may not be useful as the objectives for the design, optimization, and calibration of ISPs in terms of color reproduction.

\subsection{CD Formulae}
The research of CD in an attempt to answer the question of ``how different are two colors perceived'' can be dated back to researchers such as Helmholtz and Schr\"{o}dinger~\cite{kuhni2003historical}, Wright and Pitt~\cite{wright1934hue}, and later MacAdam~\cite{macadam1942visual}, who introduced the famous MacAdam ellipse, a region on a chromaticity diagram which contains all visually indistinguishable colors. CD formulae are widely applicable to textile, illumination, photography, television, and printing industries~\cite{luo2001ciede2000}. Up to date, more than $40$ CD measures~\cite{jaramillo2019evaluation} have been proposed, and the CIE has recommended seven of them (in chronological order): CIEUVW~\cite{wyszecki1963cieuvw}, CIELAB~\cite{robertson1977cielab}, CIELUV~\cite{robertson1977cielab}, CIE94~\cite{mcdonald1995cie94}, CIEDE2000~\cite{luo2001ciede2000}, CIECAM02~\cite{luo2013ciecam02}, and CIECAM16~\cite{li2016ciecam16}. CIELAB and CIELUV were jointly recommended in 1976 for the surface color and the TV/illumination industries, respectively. Since its debut, CIELAB~\cite{robertson1977cielab} was the most successful CD formula, widely adopted in practical applications. However, CIELAB was calibrated and tested using limited experimental data. In the 1980s, a series of experiments carried out by Luo and Rigg~\cite{luo1986bfdp}, Alman \etal~\cite{alman1989performance}, and Berns \etal~\cite{berns1991rit-dupont} demonstrated that the CIELAB color space was not as perceptually uniform as intended. On top of CIELAB, CIE94~\cite{mcdonald1995cie94} was derived and recommended in 1994. CIE94 incorporates application-specific weights and parametric factors to address the non-uniformity of CIELAB and to handle different illumination/viewing conditions. Three datasets were used in the development of CIE94, namely RIT-DuPont~\cite{berns1991rit-dupont}, Witt~\cite{witt1999witt}, and BFD-P~\cite{luo1986bfdp}. Nevertheless, CIE94 still did not adequately resolve the perceptual uniformity issue. In 2001, the joint ISO/CIE standard CIEDE2000~\cite{luo2001ciede2000} was published, which is considerably more sophisticated and computationally involved than its predecessors CIELAB and CIE94. CIEDE2000 applied a total of five corrections with respect to hue rotation and compensations for neutral colors, lightness, chroma, and hue~\cite{luo2001ciede2000}. It was constrained using the combined weighted dataset (COM) (see Section~\ref{sec:database}). {However, the aforementioned CIELAB-based formulae only work
under a set of pre-defined viewing conditions, \eg, D65 illumination, mid-grey background, and hairline separation~\cite{luo2013ciecam02}, limiting their usage in color management systems. Later, Luo \etal~\cite{luo2013ciecam02} proposed CIECAM02, a color appearance model that can predict color appearances concerning viewing conditions using separate chromatic and luminance adaptations. CIECAM16~\cite{li2016ciecam16} further simplified the CIECAM02 by merging the two adaptations without losing accuracy on color datasets.} Apart from CIE-recommended formulae, CMC~\cite{bsi1998cmc}, BFD~\cite{luo1987bfd}, HyAB/HyCH~\cite{abasi2020distance}, and J$_z$a$_z$b$_z$~\cite{safdar2017jzazbz}, are also frequently used in different contexts.

The aforementioned metrics are mostly CIELAB-based and intended for measuring CDs of homogeneous color patches.
When it comes to natural photographic images, it seems straightforward to define the CD of two images of the same scene as the average of CDs over corresponding pixels~\cite{hong2006new}. However, this notion of CD is different from how humans make color sensation of natural photographic images, where homogeneous textured regions are preferentially attended and compared within a larger spatial context~\cite{zhang1997scielab,jaramillo2019evaluation}. To incorporate spatial context into CD assessment, one line of research is to compute the weighted sum of pixel/patch CDs. Zhang and Wandell~\cite{zhang1997scielab} proposed a spatial extension of CIELAB (S-CIELAB), which adds a spatial low-pass filtering as pre-processing. Johnson and Fairchild~\cite{johnson2003top} and Choudhury \etal~\cite{choudhury2021image} proposed other S-CIELAB variants, by applying human contrast sensitivity functions to create pattern-color separable filters. Hong and Luo~\cite{hong2006new} assigned higher weights to spatially homogeneous regions that occupy larger areas or have larger predicted CDs. General-purpose image quality models, \eg, PSNR, SSIM~\cite{wang2004image}, VSI~\cite{zhang2014vsi}, PieAPP~\cite{prashnani2018pieapp}, LPIPS~\cite{zhang2018lpips}, and DISTS~\cite{ding2020image}, treat CDs as a form of ``visual distortions'' to compute perceived image differences. Nevertheless, nearly all CD measures have been evaluated on small datasets with a very limited number of visual examples, and thus their generalization to smartphone photographs is not clear. 
\begin{figure*}[t]
\centering
\begin{minipage}[t]{1.0\textwidth}
    \centering
    \subfloat[]{\includegraphics[width=0.30\textwidth]{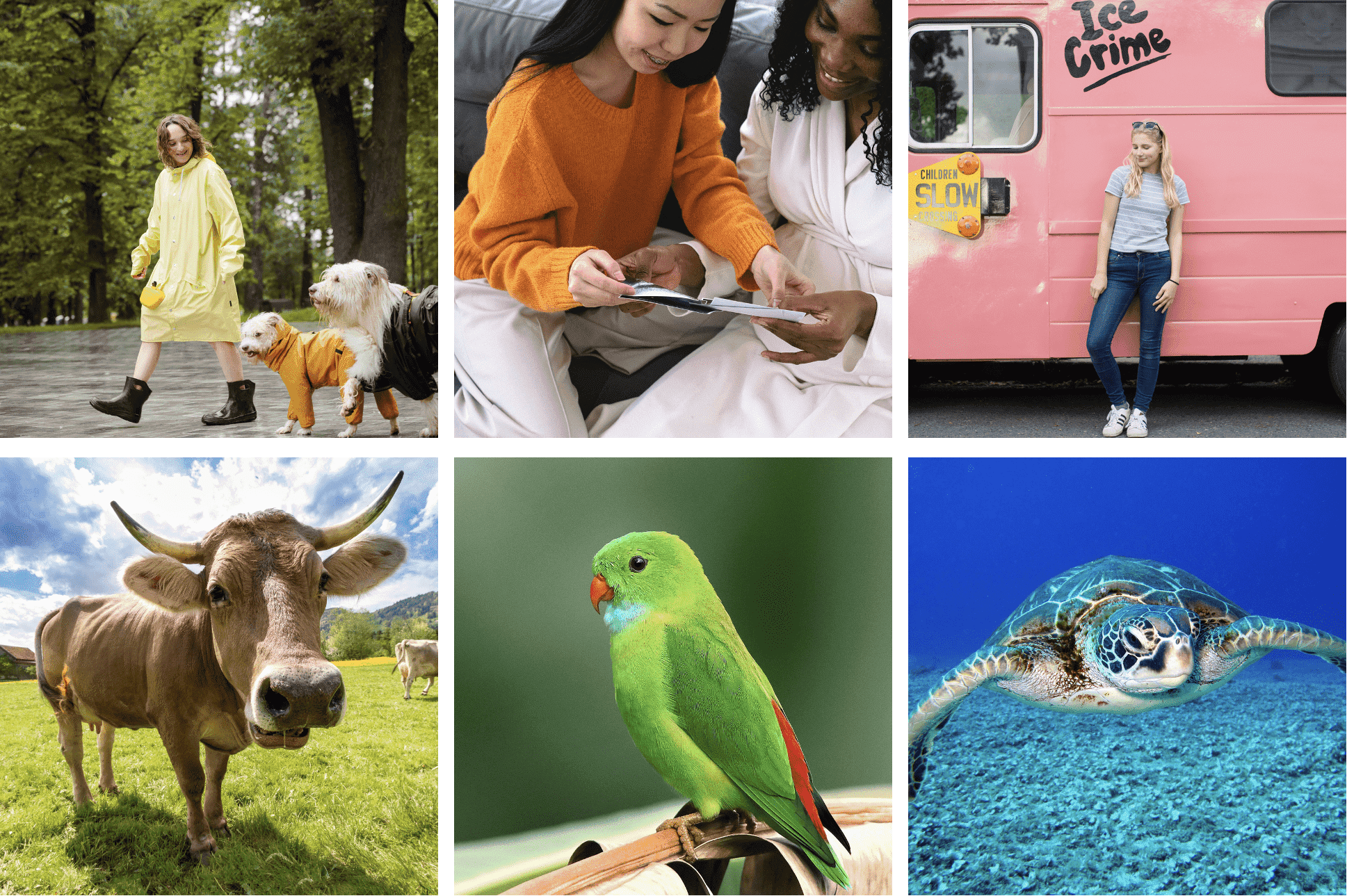}}\hskip.8em
    \subfloat[]{\includegraphics[width=0.30\textwidth]{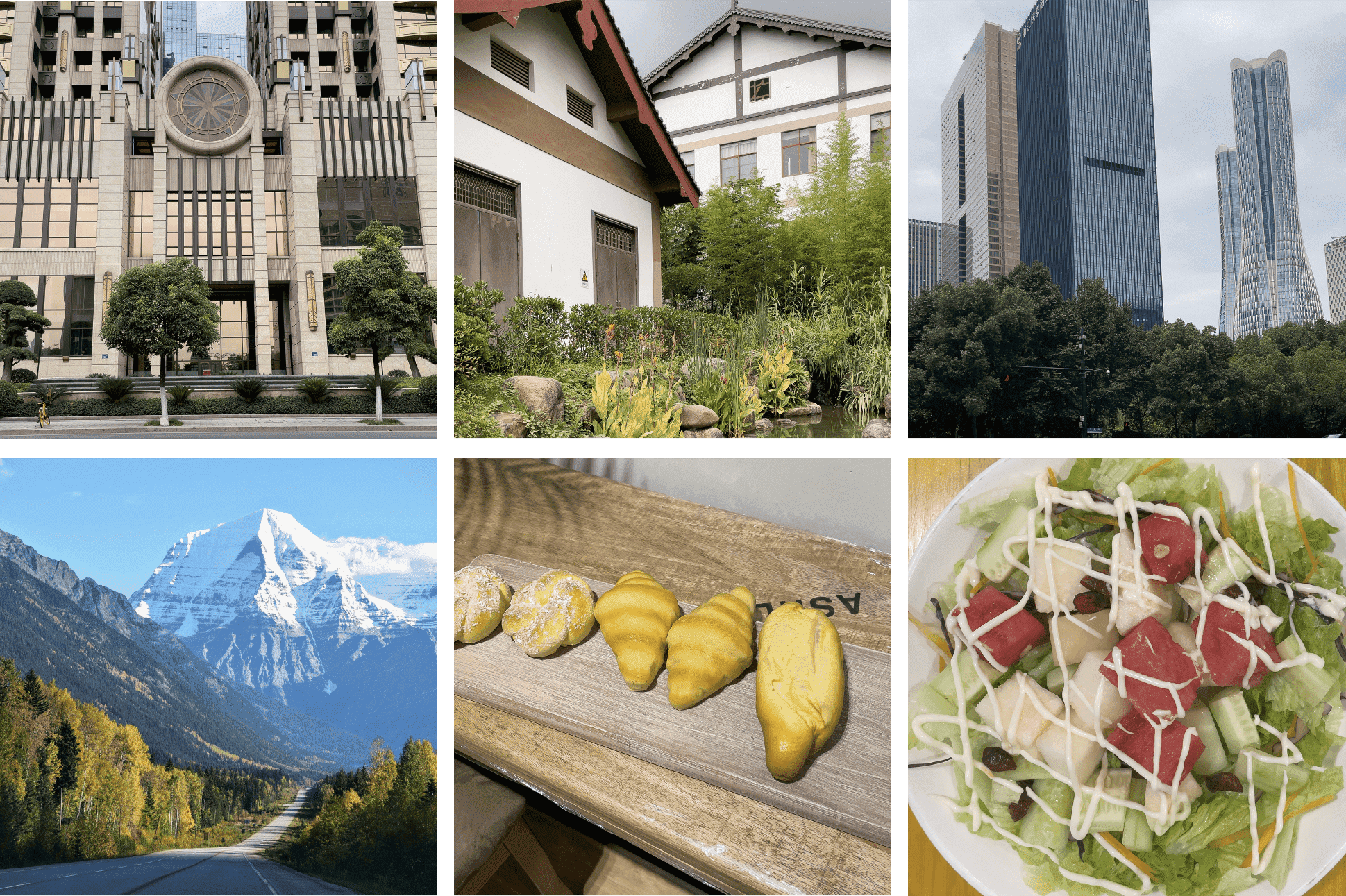}}\hskip.8em
    \subfloat[]{\includegraphics[width=0.30\textwidth]{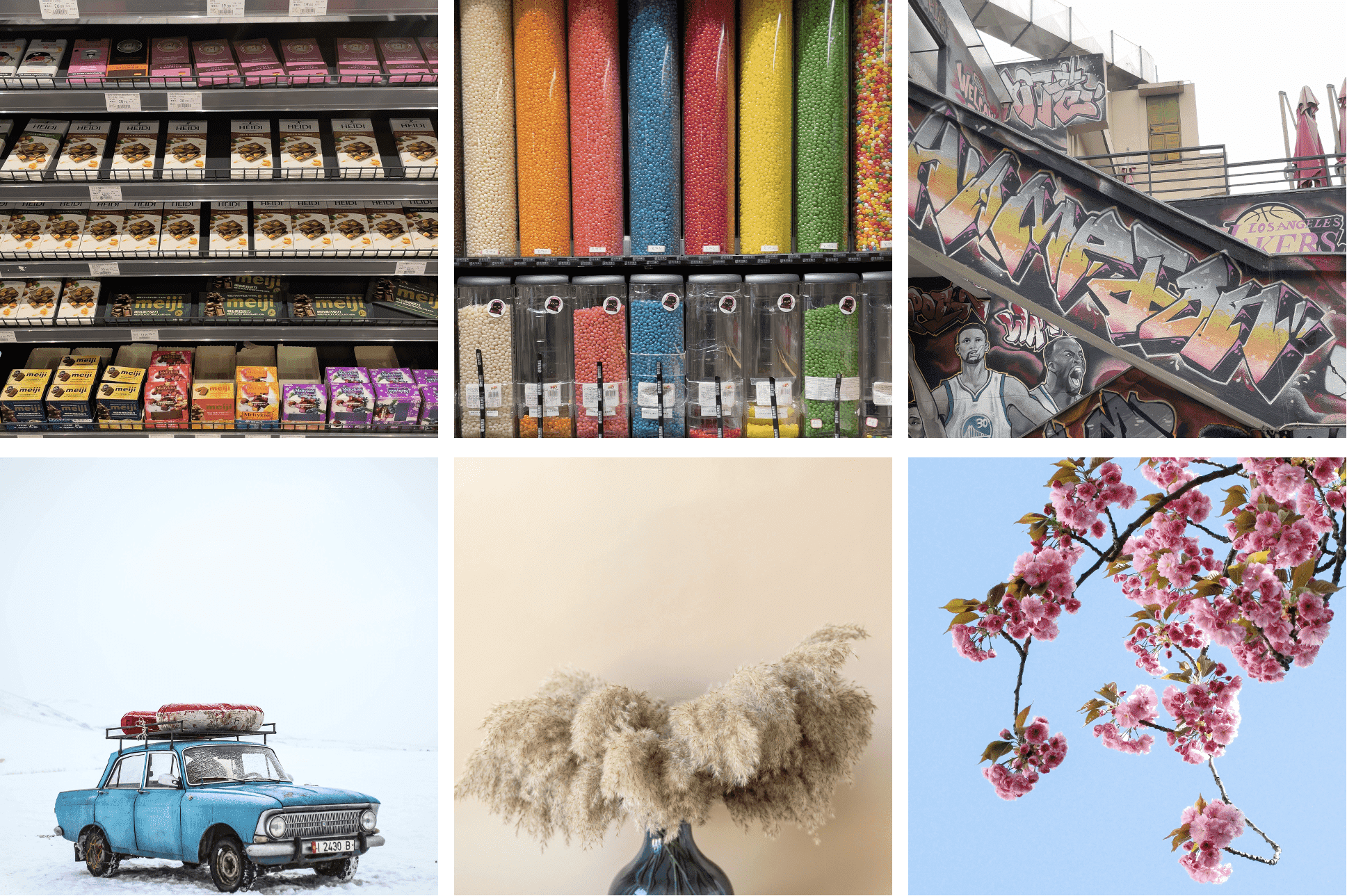}}
\end{minipage}
\begin{minipage}[t]{1.0\textwidth}
    \centering
    \vspace{.2em}
    \subfloat[]{\includegraphics[width=0.30\textwidth]{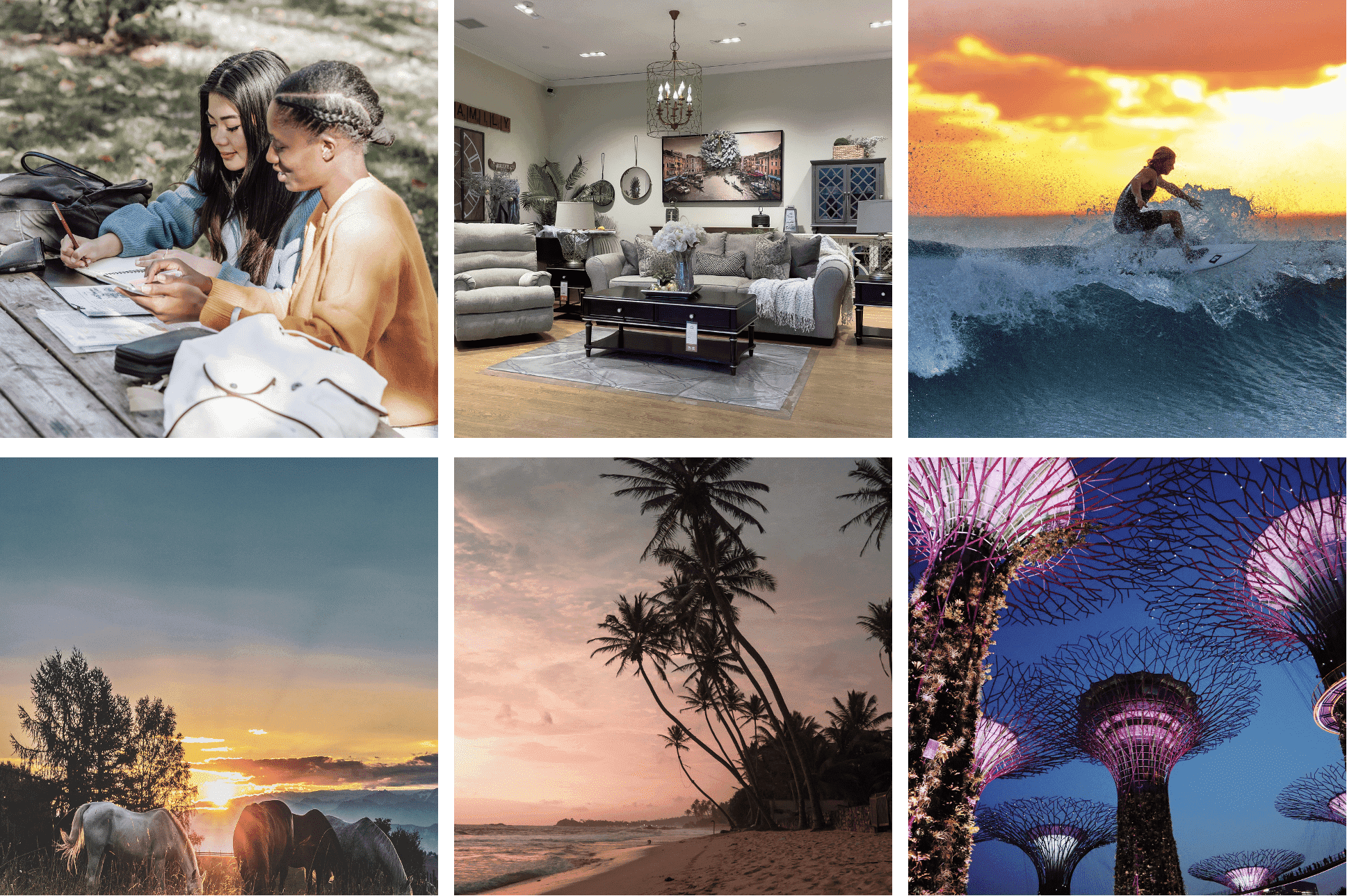}}\hskip.8em 
    \subfloat[]{\includegraphics[width=0.30\textwidth]{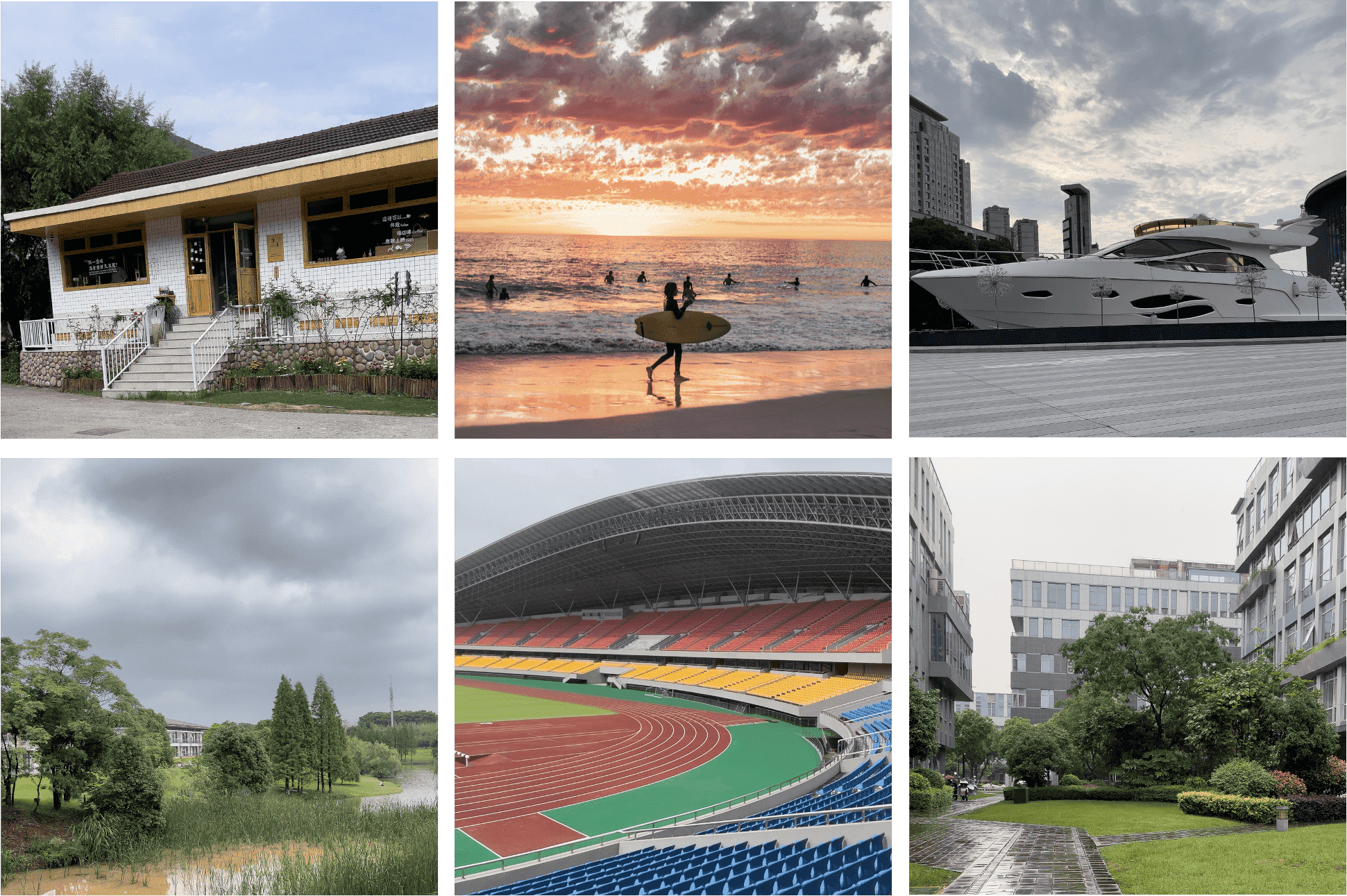}}\hskip.8em 
    \subfloat[]{\includegraphics[width=0.30\textwidth]{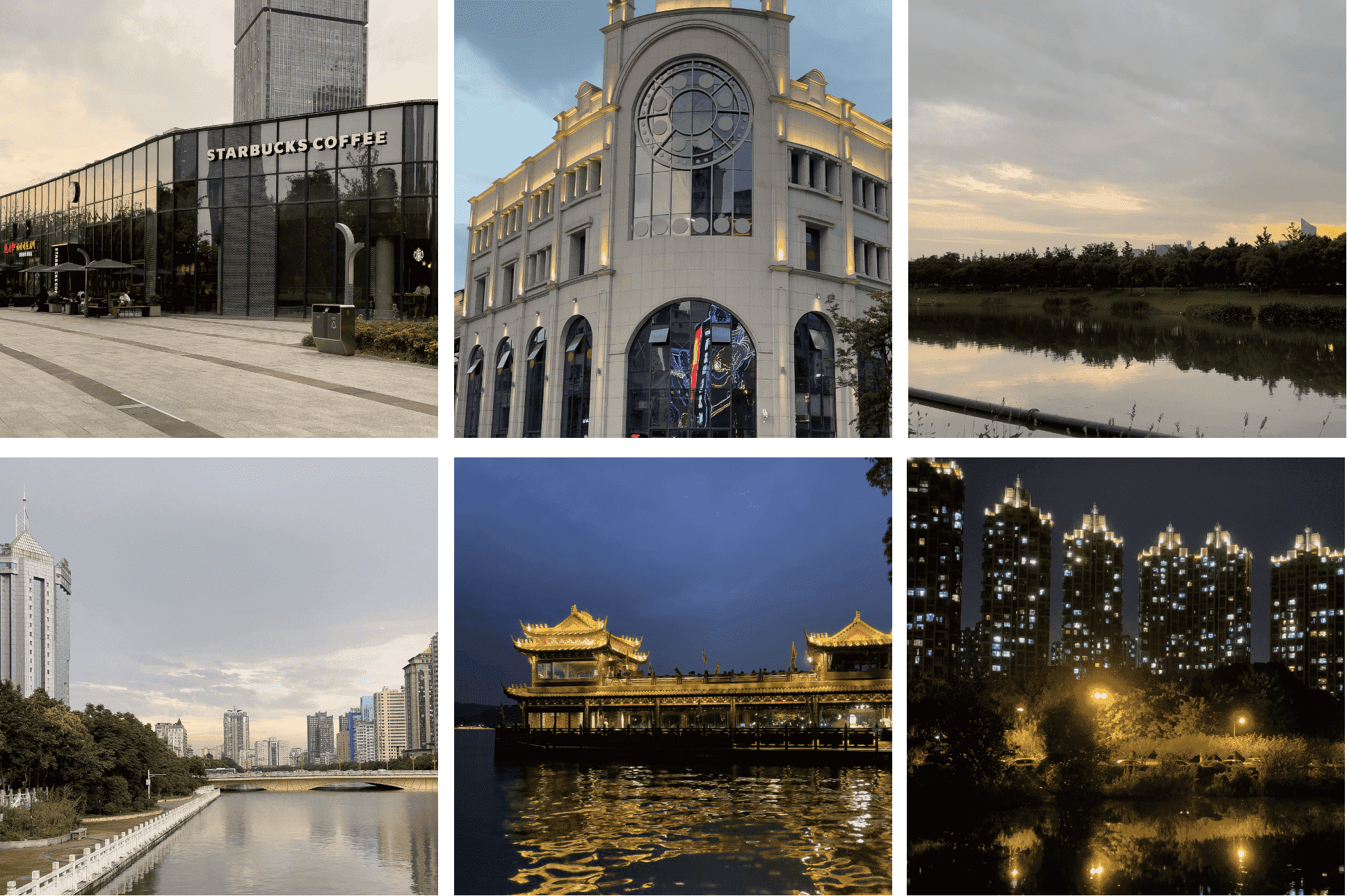}}
\end{minipage}
\caption{Representative images from the proposed SPCD dataset. (a) Human and animal. (b) Cityscape, landscape, and food. (c) Background complexity. (d) Lighting condition. (e) Weather condition. (f) Camera mode.}
\label{fig:cd_map}
\end{figure*}
\subsection{CD Datasets}
\label{sec:database} The datasets for the development of CD measures can be categorized into two classes: homogeneous color patch datasets and natural image datasets~\cite{liu2013color}.
The CIE-recommended CD metrics were developed using homogeneous color patch datasets, \eg, CIELAB on the Munsell dataset~\cite{newhall1943final}, and  CIEDE2000~\cite{luo2001ciede2000} on the COM dataset (\ie, the weighted combination of the BFD-P~\cite{luo2001ciede2000}, Leeds~\cite{kim1997leeds}, Witt~\cite{witt1999witt}, and RIT-DuPont~\cite{berns1991rit-dupont} datasets).

To study, develop, and recommend CD metrics for natural images, the CIE established the \textit{technical committee (TC) 8-02 color difference evaluation for images}~\cite{jaramillo2019evaluation}, which was closed in $2001$ with a technical report. This report lists a series of psychophysical experiments on natural color images, with the goal of calibrating CD formulae by finding perceptual colorimetric tolerances on color images. However, the number of test images is too small in these experiments to be sufficiently representative of the natural image manifold. Moreover, the CDs between images are induced by linear and quasi-linear color transforms, which are certainly over-simplifications in the age of smartphone photography. For example, Stokes~\cite{stokes1991colorimetric} used six reference images manipulated by ten (quasi-)linear functions in the CIELAB dimensions. Song and Luo~\cite{song2000testing} used four reference images, which were systematically rendered by varying lightness, chroma, mixed lightness and chroma, and hue on the CRT display. Zhang and Wandell~\cite{zhang1997scielab} reported the testing results of S-CIELAB merely on a single JEPG compressed image. Uroz \etal~\cite{uroz2002perception} tested four printed reference images with systematic transforms (\ie, pixel-wise power functions used in the printing system) and random color changes. Gibson \etal~\cite{gibson2000colorimetric} conducted colorimetric tolerance experiments on four displays (two LCDs, one CRT, and one hardcopy) of three images altered via three simple transfer curves. Liu \etal~\cite{liu2013color} collected a CD dataset 
with $100$ images ($5$ references $\times$ $20$ alternations) annotated using the categorical judgment method to optimize the parametric factors in CIELAB-based CD formulae. In addition, Jaramillo \etal~\cite{jaramillo2019evaluation}, Lee \etal~\cite{lee2014towards}, Prashnani \etal~\cite{prashnani2018pieapp}, and Zhang \etal~\cite{zhang2018lpips} used a color-related subset from TID2013~\cite{jaramillo2019evaluation} as an evaluation set. 

Despite the demonstrated effort, existing natural photographic image datasets for CD assessment are small and over-simplified, many of which are private. As a consequence, due to the lack of large-scale, human-rated, and publicly available datasets,
existing CD measures have not been rigorously compared in terms of their abilities to predict perceptual CDs of smartphone photographs.

\section{SPCD Dataset}
In this section, we create so far the largest natural image dataset, namely SPCD, tailored for perceptual CD assessment of smartphone photographs.
We first present the construction of SPCD, with emphasis on the realistic sources of CDs. We then describe the psychophysical experiment in detail, including the environmental setup and subjective testing, followed by subjective data processing.

\subsection{Dataset Construction}\label{subsec:dcon}
\noindent\textbf{Image Selection.} We gather a total of $15,335$ color images out of $1,000$ distinct natural scenes, among which $667 \times 6 = 4,002$ are from $667$ scenes captured by the authors with six smartphones, $333$ (one image per scene) are downloaded from the Internet that carry Creative Commons licenses, and the remaining $1,000 \times 11 = 11,000$ are color alternations of $1,000$ distinct scenes. In accordance with the DXOMARK's tests~\cite{dxomark}, the natural scenes are selected to span a variety of realistic shooting scenarios in terms of
\begin{itemize}
    \item \textbf{content diversity}: animal, plant, human, food, landscape, and cityscape;
    \item \textbf{background complexity}: cluttered and single-color;
    \item \textbf{lighting condition}: diffuse light, front light, back light, natural light in the sunrise, noon and sunset, and night;
    \item \textbf{weather condition}: {sunshine}, cloudy, and rainy;
    \item \textbf{camera mode}: HDR and night.
\end{itemize}
Fig.~\ref{fig:cd_map} shows representative images from SPCD. It is noteworthy that all images are resized and cropped to $1,024\times1,024$, and stored in uncompressed format. 
\vspace{0.4em}\\
\begin{figure}
    \centering
    \begin{minipage}[t]{0.4\textwidth}
        \subfloat[]{\includegraphics[width=1\textwidth]{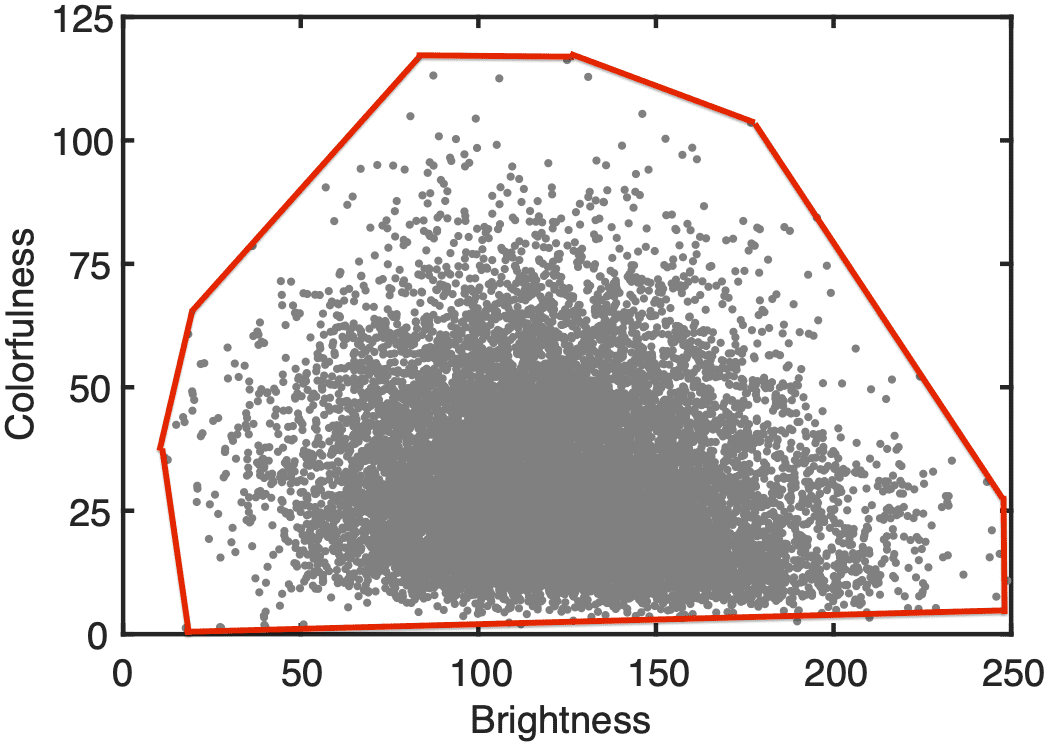}}
        \hskip.8em
        \subfloat[]{\includegraphics[width=1\textwidth]{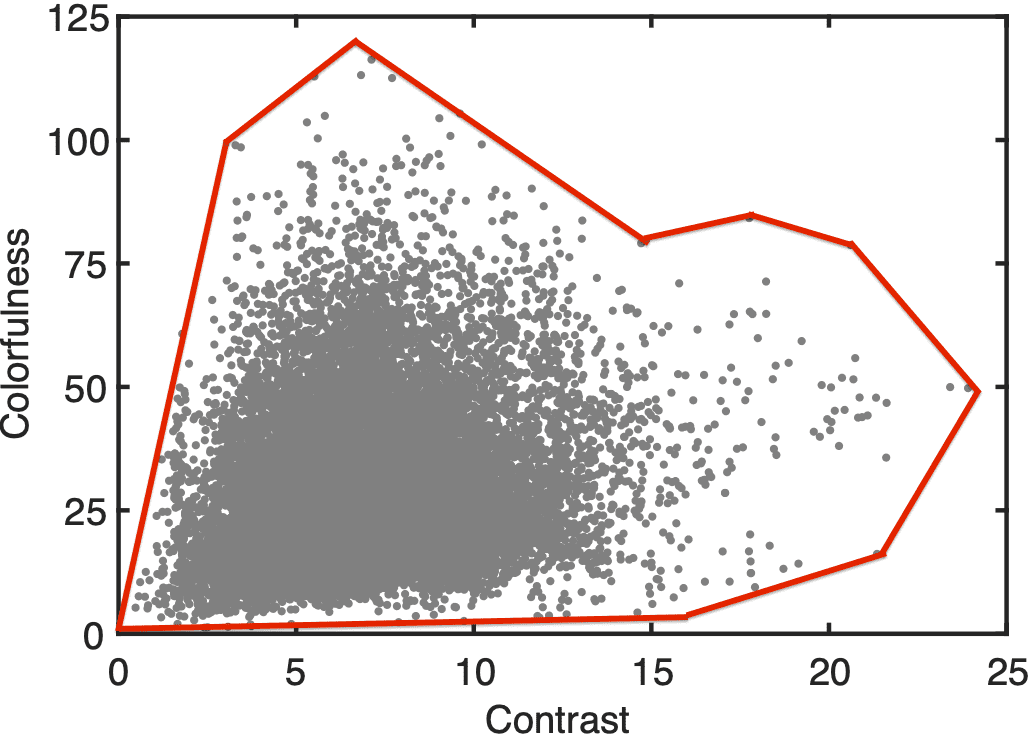}}
    \end{minipage}
    \caption{Pairwise feature distributions with the corresponding convex hulls of SPCD. (a) Brightness against Colorfulness. (b) Contrast against Colorfulness.}
    \label{fig:attribute_distribution}
\end{figure}
\noindent\textbf{CD Generation}. We generate four types of CDs that are naturally occurring in smartphone photography.
\begin{itemize}
  \item[\rom{1}] Same scene captured by different smartphones. We use six flagship smartphones - Apple iPhone 12 Pro, HUAWEI Mate40 Pro, OnePlus 7 Pro, Samsung S21Ultra, OPPO Find X3 Pro, and Xiaomi 11 Ultra.
  Since the camera system and the associated ISP are proprietary, and vary among different smartphone brands, the captured pictures inevitably exhibit different color appearances, especially in night scenes. One subtlety is that different cameras may produce images of different sizes and displacements, which require cropping and alignment. We adopt a simple feature-based method\footnote{\url{https://github.com/khufkens/align\_images.}} to estimate an affine matrix for global registration. There are indeed more sophisticated methods to provide a better account of misregistration. We intentionally opt not to do this, so we can test the robustness of CD measures to this imperceptible level of misregistration to the human eye\footnote{In the current work, we discard pictures that fail to be registered to the imperceptible level.}. 
  \item[\rom{2}] Same image altered by Photoshop to simulate the effect of ISP functions. Since white balance, color correction, tone mapping, and gamma correction are the four main sub-modules that are highly related to color reproduction and manipulation, we synthesize these color transforms by adjusting the corresponding parameters in Photoshop\footnote{More specifically, white balance adjusts color temperature and tint; color correction changes the RGB curves separately; tone mapping tunes exposure; and gamma correction alters mid tone range by dragging the gamma slider.}. 
  \item[\rom{3}] Same image post-processed by built-in filters of the iPhone. We select nine filters to produce different artistic styles: 1) \textit{vivid}, which increases the photo’s contrast, 2) \textit{vivid warm}, being \textit{vivid} while overlaying warm tones, 3) \textit{vivid cool}, being \textit{vivid} while overlaying cool tones, 4) \textit{dramatic}, which increases the shadows and decreases highlights, 5) \textit{dramatic warm}, being \textit{dramatic} while overlaying warm tones, 6) \textit{dramatic cool}, being \textit{dramatic} while overlaying cool tones, 7) \textit{mono}, which increases the sharpness, 8) \textit{silver tone}, which increases the shadows, and 9) \textit{noir}, which increases the contrast.
  \item[\rom{4}] Same image reproduced with incorrect ICC profiles. This may be the primary reason when a color management system fails to maintain the color appearance of a photographic image across media devices. For example, an sRGB image may look over-saturated on a monitor that supports a wider color gamut, \eg, DCI-P3~\cite{apple2017DisplayP3} and Rec. 2020~\cite{itu2020parameter}. We simulate two cases: sRGB images are mis-displayed in DCI-P3\footnote{DCI-P3 images are first converted to XYZ, subsequently to sRGB, and finally saved in the sRGB mode. The display is set to DCI-P3 mode. Visually, those photographs would appear more chromatic than correctly shown.}, and vice versa.
\end{itemize}

\begin{figure}
    \centering 
    \includegraphics[width=1\columnwidth]{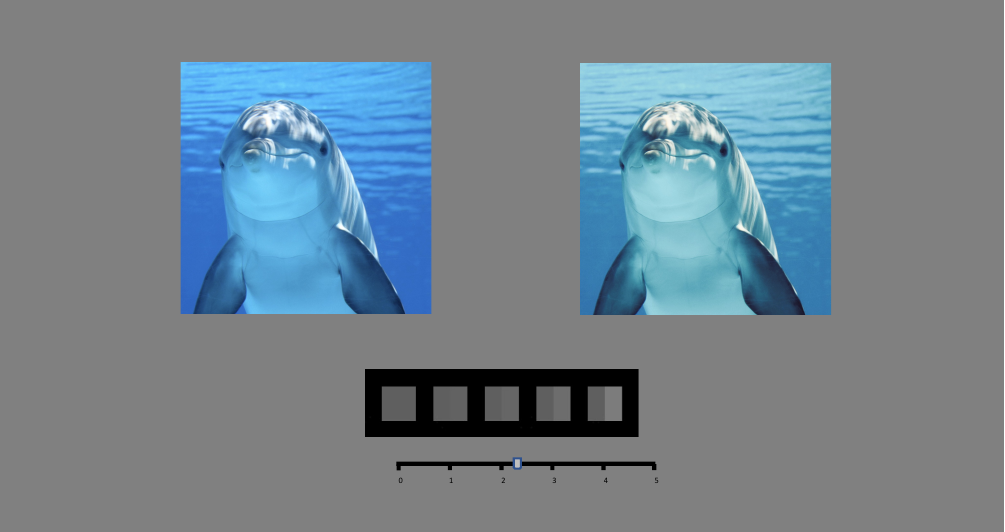}
    \caption{The graphical user interface for subjective testing. } 
    \label{fig:interface}
\end{figure}

As suggested by Winkler~\cite{winkler2012analysis}, we quantify the diversity of SPCD using three computational measures: colorfulness (with a larger value indicating more colorfulness)~\cite{hasler2003measuring}, brightness (with a larger value indicating more brightness), and contrast (with a larger value indicating higher contrast)~\cite{matkovic2005global}. Fig.~\ref{fig:attribute_distribution} plots the convex hulls of brightness versus colorfulness and contrast versus colorfulness, verifying the relative uniformity and wide coverage. We then randomly sample $10,005$ non-perfectly aligned image pairs with CDs of Type \rom{1}, and $19,995$ perfectly aligned image pairs with CDs induced collectively by color transforms of Types \rom{2}, \rom{3}, and \rom{4}, resulting in a total of $30,000$ image pairs in SPCD.
 
\subsection{Psychophysical Experiment}
Unlike perceptual image quality~\cite{wang2006modern}, which can be assessed in a relatively uncontrolled online crowdsourcing platform with reasonable consistency among subjects, collecting reliable human judgments of perceptual CDs would rather require a carefully controlled laboratory environment because the color perception depends highly on the accuracy of the display for color reproduction and the ambient environment in which the display is placed.
\vspace{0.4em} \\
\noindent\textbf{Experimental Setup.} The subjective testing environment is setup in a completely dark indoor office with no illumination and little reflection. A customized graphical user interface is devised for perceptual CD collection. As shown in Fig.~\ref{fig:interface}, the background is set to be neutral gray. A pair of images with the same content but different color appearances are displayed in full resolution, whose CD is rated with reference to five grayscale sample pairs. Such a method for CD assessment was originated from the textile industry for evaluating color fastness~\cite{standard2002textiles}, was first used for academic research by Luo and Rigg~\cite{luo1987bfd}, and was proved by later studies~\cite{xu21testing} as a reliable means of collecting CDs. As suggested in~\cite{standard2002textiles}, the lightness differences of the five grayscale pairs in the CIELAB unit (\ie, $\varDelta E_{ab}^*$) are around $0$, $1.7$, $3.4$, $6.8$, and $13.6$, respectively. The corresponding actually measured values ($\varDelta E_{ab}^m$) by a tele-spectroradiometer are listed in Table~\ref{tab:measure}, which are considered acceptably close to the recommended values. 
A scale-and-slider applet is located at the bottom to collect \textit{continuous} CD scores, with reference to the sample grayscale pairs. Subjects are given unlimited time to rate one pair with a minimum of three seconds. The viewing distance is fixed to one meter. Ten male and ten female observers, who have normal color vision and normal or corrected-to-normal visual acuity, participate in the subjective experiment, where the color normal vision is evaluated using the Ishihara's Color Vision Test. Each subject is asked to give CD scores to all image pairs. To reduce the fatigue effect, the subjects are required to take a break after a $30$-minute experiment. Each subject completed $300$ pairs per day, and the entire experiment lasts a total of four months. In total, we collect $20 \times 30,000 = 600,000$ CD scores.
\vspace{0.4em} \\

\begin{table} [tbp]
	\caption{The conversion between the grayscale grade levels and the perceptual CDs, measured by a tele-spectroradiometer and predicted by Eq. \eqref{eq:convertion}}
	\label{tab:measure}
	\begin{center}
	\begin{tabular}{l|ccccc}
	    \toprule
		Grayscale Pair &  GS0 & GS1 & GS2 & GS3 & GS4\\
		\hline
	    Grade Level & 0 & 1 & 2 & 3 & 4 \\
		Measured $\varDelta E_{ab}^{m}$ & 0.00 & 1.83 & 3.59 & 6.45 &12.66 \\
		Predicted $\varDelta V$ & 0.31 & 1.46 & 3.42 & 6.79 & 12.56 \\
		\bottomrule
	\end{tabular}
\end{center}
\end{table}

\begin{figure}
    \centering
    \hskip-.8em
   \includegraphics[width=0.8\columnwidth]{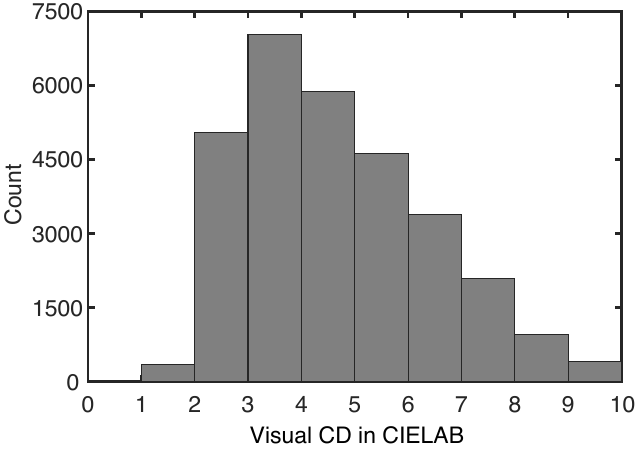}
    \caption{Empirical distributions of $30,000$ perceptual CDs in SPCD.}
    \label{fig:rating_distribution}
\end{figure}

\noindent\textbf{Display Characterization}.
\label{sec:display_characterization}
Two EIZO CG319X 31.1" LCD monitors are adopted in the experiment, with a spatial resolution of $4,096\times2,160$ pixels and a maximum contrast ratio of $1,500:1$. It is one of the state-of-the-art and award-winning displays with HDR and wide color gamut (WCG) capabilities. The display peak white is set to be $100$ cd/m$^2$ at D$65$. We conduct comprehensive experiments to characterize the display's colorimetric characteristics~\cite{kwak2000characterisation}, including
\begin{itemize}
\item \textbf{temporal stability:} testing the short-term and mid-term stabilities of peak white luminance after a cold start;
\item \textbf{spatial independence:} assessing the dependency of background via measuring the color difference of central gray color patches on various bright and colorful backgrounds; 
\item \textbf{spatial uniformity:} analyzing the spatial color shift by measuring the difference of the peak white among different positions;
\item \textbf{channel independence:} checking the colorimetric additivity and interactivity of the three tristimulus channels;
\item \textbf{chromaticity constancy:} characterizing the locus of chromaticity changes of primary and achromatic colors with respect to the digital input values of each channel;
\item \textbf{color gamut:} measuring the reproducible color range.
\end{itemize}

We use the CIE-recommended gamma-offset-gain (GOG) display model~\cite{berns1996methods} to nonlinearly relate the digital input and the luminance of each channel in RGB color space. We adopt the tele-spectroradiometer - JETI Specbos 1211uv, whose accuracy is within $2\%$ when measuring Illuminant A of $100$ cd/m$^2$, under the assumption of the CIE 1964 standard colorimetric observer~\cite{edition2004colorimetry}.
The obtained model is tested to have a performance of $0.56~\varDelta E_{ab}^*$ based on the Macbeth ColorChecker Chart, indicating that our display is suitable for color-related vision experiments.

\begin{table}[t]
    \caption{Min, max, median and mean STRESS, SRCC, and PLCC between two randomized subgroups with equal size across $100$ splits}
    \label{tab:internal_consistency}
    \vspace{-.3cm}
	\begin{center}
		\begin{tabular}{l|cccc}
    		\toprule[1pt]
			Criterion & Min & Max & Median & Mean\\
			\hline
			STRESS$\downarrow$ & $17.092$ & $23.702$ & $18.750$ & $19.263$\\
			SRCC$\uparrow$ & $0.823$ & $0.887$ & $0.866$ & $0.864$ \\
 		PLCC$\uparrow$ & $0.819$ & $0.890$ & $0.869$ & $0.866$\\
			\bottomrule[1pt]
		\end{tabular}
	\end{center}
\end{table}

\subsection{Subjective Data Processing}
\noindent \textbf{Transform from Raw Scores to Perceptual CDs}.
Following~\cite{coates1972measurement}, we transform raw subjective scores given in the grayscale grade unit to the CIELAB unit, $\varDelta E_{ab}^*$ (\ie, the perceptual CD). As suggested by the ISO standard~\cite{standard2002textiles}, we fit an exponential function, $a\exp(bG)+c$, for this conversion, where $G\in\{0, 1, 2, 3, 4\}$ denotes the grade levels and $\{a,b,c\}$ are parameters to be fitted. Through minimization of the differences between the measured CDs ($\varDelta E_{ab}^m $) and corresponding predicted CDs ($\varDelta V$), we obtain the fitted exponential function:
\begin{align} 
\label{eq:convertion}
\varDelta V = 1.6036\exp({0.5391G})-1.2943.
\end{align}
As shown in Table~\ref{tab:measure}, the measured CDs can be well predicted by Eq.~(\ref{eq:convertion}), with the largest prediction error less than $0.37$ $\varDelta E_{ab}^*$. This is far below the just noticeable CD of a normal observer, indicating that the conversion is reasonable. 
\vspace{0.4em} \\
\noindent\textbf{Outlier Detection and Subject Rejection}. We follow the recommendation in~\cite{bt2002methodology} to detect outlier ratings and reject invalid subjects. Specifically, for each image pair, we first convert the raw subjective scores to perceptual CDs, which are identified as outliers if they lie out of three standard deviations. Subjects with an outlier rate $\le5\%$ are considered valid. After data purification, we find that all subjects are valid, and $1.09\%$ of the ratings are outliers and are subsequently removed. The mean of valid CDs for each image pair is treated as the ground-truth perceptual CD~\cite{bt2002methodology}.
\vspace{0.4em} \\
\noindent\textbf{Results and Analysis.} Fig.~\ref{fig:rating_distribution} plots the histogram of perceptual CDs of $30,000$ image pairs in SPCD, which is well fitted by a unimodal distribution with mode around $3.5$. To verify the reliability of the collected CDs, we randomly split the subjects into two subgroups of equal size, and compute the standardized residual sum of squares (STRESS)~\cite{garcia2007measurement}, the Spearman’s rank correlation coefficient (SRCC), and the Pearson linear correlation coefficient (PLCC) (see Section~\ref{sec:cd-net}) between their respective mean perceptual CDs. We repeat this procedure $100$ times, and show the results in Table~\ref{tab:internal_consistency}, where high consistency between two subgroups has been observed.

\begin{figure}[t]
\centering
    \includegraphics[width=1\columnwidth]{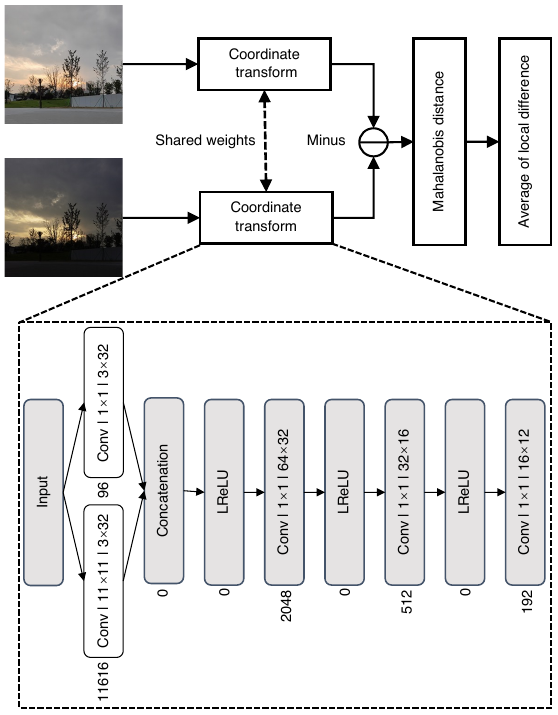}
    \caption{System diagram of CD-Net for perceptual CD assessment. The parameterization of convolution is denoted as ``filter | kernel size | input channel $\times$ output channel ''. The number of trainable parameters for each layer is given at the bottom, yielding a total of $14,464$.}
    \label{fig:network}
\end{figure}

\section{Proposed CD-Net}
In this section, we aim to learn an end-to-end CD measure, which generalizes several previous CD formulae. At a high level, our method, called CD-Net, consists of a lightweight DNN for coordinate transform, a Mahalanobis distance metric for local CD calculation, and a global average operation for overall CD assessment.

\subsection{Formulation}
Conventional CD measures generally consist of two steps: coordinate transform (to a ``perceptually uniform'' color space) and distance calculation~\cite{ luo2001ciede2000, imai2001perceptual, jaramillo2019evaluation}. Assume a training dataset $\mathcal{D} = \{(x^{(i)}, y^{(i)}), \varDelta V^{(i)}\}_{i=1}^{M}$, where $(x^{(i)}, y^{(i)})$ is the $i$-th image pair, $\varDelta V^{(i)}$ is the corresponding ground-truth perceptual CD, 
and $M$ is the number of image pairs. Our goal is to learn a differentiable parametric distance function $\varDelta E: \mathbb{R}^{N\times3} \times \mathbb{R}^{N\times3} \mapsto \mathbb{R}$, which takes two $N$-dimensional RGB images of the same scene as input, and computes a nonnegative scalar to represent the predicted CD. We design CD-Net as a conceptually generalized measure of conventional CD formulae, which also consists of the same two operations.
\begin{itemize}
    \item [1)] \textbf{Coordinate Transform}. Natural photographic images are generally stored, rendered, and reproduced in RGB color space, which is not perceptually uniform. Thus, the very first step of CD-Net is to transform an RGB pixel into a new vector space of possibly more than three dimensions. Since humans make color sensation of a local image region by comparing it within a larger spatial context~\cite{ zhang1997scielab,jaramillo2019evaluation}, the desired coordinate transform should take the neighboring pixels into account. 
    \item [2)] \textbf{CD Calculation}. This step computes the CD between two points in the transformed space. It is desirable for this computation to satisfy several mathematical properties such as non-negativity, symmetry, the identity of indiscernibles, and triangle inequality, giving rise to a mathematical metric in the transformed space and even in RGB space (see Section~\ref{subsec:ft}). In practice, one usually seeks a single overall CD score between the two photographic images. Thus, local CD measurements should be aggregated into a global CD score~\cite{hong2006new}.   
\end{itemize}

\subsection{Coordinate Transform}
The top panel of Fig.~\ref{fig:network} illustrates the system diagram of CD-Net. Given an RGB image $x \in \mathbb{R}^{H \times W \times 3}$, where $H$ and $W$ are the spatial height and width, respectively, we feed it to a lightweight DNN, $f_\theta$ parameterized by a vector $\theta$, for coordinate transform. To incorporate spatial context, we take inspiration from prior work of multi-branch DNNs~\cite{szegedy2016inceptionnet}, and use a front-end filter bank with convolutions of $T$ different kernel sizes for multi-scale processing:
\begin{align}\label{eq:standardconv}
z^{(t)}_{p,q,r} = \sum_{(i,j) \in \mathcal{N}_t} \sum_{k=1}^3 w_{i,j,k,r} x_{p+i, q+j,k},
\end{align}
where $\mathcal{N}_t$ is the neighboring grid of the $t$-th convolutions, $(p,q)$ denotes the center location, and $r$ indexes the convolution channel. We then concatenate the multi-scale filter responses:
\begin{align}\label{eq:concat}
z = \mathrm{concat}\left(z^{(1)}, z^{(2)},\ldots, z^{(T)}\right),
\end{align}
 which further undergoes several stages of $1\times1$ convolution layers with leaky ReLU as nonlinearity. The channel number of the last convolution is set to $C$. To encourage learning fine-scale local CD maps, no spatial downsampling is involved. As a result, $f_\theta(x)\in \mathbb{R}^{H\times W\times C}$ has the same spatial size of $x$, manifesting itself as a coordinate transform from the three-dimensional RGB space to a $C$-dimensional feature space with possibly improved perceptual uniformity.
 
 Conceptually, the proposed coordinate transform $f_\theta$ in CD-Net can be seen as a generalization of color space transforms used in conventional CD formulae. This is because as a universal approximator, $f_\theta$  may easily fit the piece-wise nonlinear function used in CIELAB, based on which a wide range of CD measures~\cite{mcdonald1995cie94,bsi1998cmc, luo2001ciede2000} have been developed. Moreover, thanks to the front-end multi-scale filter bank, $f_\theta$ also generalizes spatial extensions of CIELAB~\cite{zhang1997scielab,ouni2008new,simone2009alternative,pedersen2012new}.

\subsection{CD Calculation}
Mahalanobis distance is one of the frequently used metrics to compute the distance between two vectors. Here, we employ it to compute the CD between two points in the transformed space~\cite{imai2001perceptual}: 
\begin{align}\label{eq:mds}
    \varDelta E(x_{ij}, y_{ij}) = \sqrt{(f(x)_{ij} - f(y)_{ij})^T S^{-1} (f(x)_{ij} - f(y)_{ij})},
\end{align}
parameterized by a learnable matrix $S \in \mathbb{S}_{+}^{C}$, where $\mathbb{S}_{+}^{C}$ is the cone of symmetric positive semi-definite real-valued $C \times C$ matrices. According to Cholesky decomposition, $S$ can be efficiently parameterized by a real lower triangular matrix $L$ with nonnegative diagonal entries, \ie, $S=LL^T$. Finally, CD-Net uses the mean of local CD measurements to evaluate the overall CD between the two images: 
\begin{align} 
    \varDelta E(x, y) = \frac{1}{HW}\sum _{i, j} \varDelta E(x_{ij}, y_{ij}).
\end{align}

The proposed CD-Net is a proper metric in the transformed space by definition. In order for CD-Net to be a proper metric in the input RGB space as well, it is necessary to enforce the \textit{injectivity} of the coordinate transform $f_\theta$: distinct three-dimensional RGB points should map to distinct $C$-dimensional feature vectors. Although the learning of invertible DNN-based transforms has experienced considerable progress, the invertibility of CNNs with multiple layers and optimized network parameters (for a particular task) is still an open research problem~\cite{ma2018invertibility}. In Section \ref{sec:verification_of_mathematical_property}, we will design experiments to empirically probe whether the proposed CD-Net satisfies the properties of a proper metric in RGB space.

Similarly in~\cite{imai2001perceptual}, the learnable Mahalanobis distance in CD-Net uses the diagonal and off-diagonal entries to include the weighting factors in CIE94~\cite{mcdonald1995cie94} and the hue rotation term in CIEDE2000~\cite{luo2001ciede2000} as special cases.
Depending on the application, it is convenient to compute a weighted summation of local CD measurements, with weightings based on error visibility~\cite{chou2007fidelity,andersson2020flip}, and region of interest~\cite{zhang2014vsi,jaramillo2019evaluation}. The histogram intersection method presented by Lee \etal~\cite{lee2005evaluation} that can compare CDs of spatially distant points is also related to our CD-Net. By replacing standard convolutions with dilated convolutions of large receptive field, CD-Net is capable of accomplishing a similar goal.

\subsection{Specification of CD-Net}
The bottom panel of Fig.~\ref{fig:network} shows the specification of the coordinate transform in CD-Net, with the design goal of being as lightweight as possible. Specifically, motivated by the parameter setting of CIE-recommended formulae~\cite{robertson1977cielab, mcdonald1995cie94, luo2001ciede2000} and SSIM~\cite{wang2004image}, we instantiate the front-end filter bank with two convolution layers of $1\times1$ and $11\times11$ kernel sizes for pixel-wise and patch-wise processing, respectively (\ie, $T=2$). After feature concatenation and rectification, we use three $1\times1$ convolution layers with LReLU in between. The number of convolution channels is initially $64$ and is reduced by 
a factor of $2$ for the subsequent two layers, and we set the channel number of the last convolution to $C=12$. The negative slope coefficient of LReLU is set to $10^{-2}$. We exclude the bias term in Eq.~\eqref{eq:standardconv} to enforce the scaling invariance of the coordinate transform~\cite{mohan2019biasfreedenosing} (\ie, $f_\theta(\alpha x) = \alpha f_\theta(x)$, for an arbitrarily fixed scalar, $\alpha$). In total, the proposed CD-Net has $14,464$ parameters in $f$ and $78$ parameters in $S$ parameterized by $L$.
  \begin{table*}[t]
    \caption{STRESS, SRCC, and PLCC between predicted CDs ($\varDelta E$) and perceptual CDs ($\varDelta V$) in SPCD. The top section lists representative CD formulae developed from homogeneous color patches. The second section contains CD measures adapted for natural photographic images. The third section includes general-purpose image quality models. The fourth section consists of JND measures. The fifth section gives trained DNNs with frequently used backbones as reference. The top two methods are highlighted in boldface}
    \label{tab:comparision}
    \vspace{-.3cm}
	\begin{center}
	    \begin{threeparttable} 
		\begin{tabular}{l|c|ccc|ccc|ccc}
    		\toprule[1pt]
			\multirow{2}*{Method} & Color & \multicolumn{3}{c|}{Perfectly aligned pairs} & \multicolumn{3}{c|}{Non-perfectly aligned pairs} & \multicolumn{3}{c}{All}\\
    		\cline{3-11}
			 & space & STRESS$\downarrow$ & PLCC$\uparrow$ & SRCC$\uparrow$ & STRESS$\downarrow$ & PLCC$\uparrow$ & SRCC$\uparrow$ & STRESS$\downarrow$ & PLCC$\uparrow$& SRCC$\uparrow$ \\
           	\hline
 		    CIELAB~\cite{robertson1977cielab} & CIELAB &$31.244$ & $0.793$ & $0.775$ & $29.639$ &  $0.690$ & $0.579$  & $31.872$ & $0.716$ & $0.666$   \\
 		    CIE94~\cite{mcdonald1995cie94} & CIELAB & $34.721$ & $0.790$ & $0.772$& $29.916$ & $0.693$ & $0.572$& $34.326$ & $0.710$ & $0.654$\\
 		    CMC~\cite{bsi1998cmc} & CIELAB & $34.113$ & $0.786$ & $0.786$& $34.125$ & $0.591$ & $0.490$& $35.936$ & $0.664$ & $0.632$\\
            CIEDE2000~\cite{luo2001ciede2000} & CIELAB & $29.975$ & $0.825$ & $0.821$& $30.347$ & $0.667$ & $0.563$& $31.439$ & $0.726$ & $0.686$\\
 		Huertas06~\cite{huertas2006performance}  & OSA-UCS &  
            $36.466$ & $0.679$ & $0.689$& $34.506$ & $0.546$ & $0.426$& $36.451$ & $0.573$ & $0.562$ \\
             HyAB~\cite{abasi2020distance}  & CIELAB &  $30.929$ & $0.787$ & $0.770$& $29.951$ & $0.677$ & $0.580$& $31.668$ & $0.714$ & $0.668$ \\
             HyCH~\cite{abasi2020distance}  & CIELAB &  $28.811$ & $0.824$ & $0.815$& $30.021$ & $0.667$ & $0.572$& $30.605$ & $0.729$ & $0.688$ \\
             CIECAM02~\cite{luo2013ciecam02}  & CIECAM02 &  $33.377$ & $0.797$ & $0.781$& $29.769$ & $0.690$ & $0.574$& $33.397$ & $0.714$ & $0.660$ \\
             CIECAM16~\cite{li2016ciecam16}  & CIECAM16  &  $31.507$ & $0.810$ & $0.799$& $29.529$ & $0.691$ & $0.577$& $32.138$ & $0.722$ & $0.673$ \\
             J$_z$a$_z$b$_z$~\cite{safdar2017jzazbz} &  J$_z$a$_z$b$_z$ &  $32.504$ & $0.779$ & $0.768$& $32.150$ & $0.640$ & $0.588$& $32.758$ & $0.701$ & $0.662$\\
 		\hline
 	        S-CIELAB\tnote{1}~\cite{zhang1997scielab} & CIELAB &  $30.094$ & $0.822$ & $0.819$& $31.804$ &  $0.631$ & $0.522$ & $32.780$ & $0.700$ & $0.657$\\
 			Imai01~\cite{imai2001perceptual} & CIELAB & $60.123$ & $0.683$ & $0.694$ & $48.573$ & $0.527$ & $0.524$ & $57.329$ & $0.597$ & $0.606$\\
 			Toet03~\cite{toet2003new} & $l\alpha\beta$& $34.941$ & $0.337$ & $0.392$& $38.624$ & $0.139$ & $0.048$& $36.216$ & $0.197$ & $0.176$\\
 			Lee05~\cite{lee2005evaluation} & CIELAB & $58.891$ & $0.734$ & $0.741$& $55.826$ & $0.622$ & $0.624$& $58.010$ & $0.697$ & $0.710$\\
 			Hong06~\cite{hong2006new}& CIELAB  & $60.557$ & $0.794$ & $0.810$& $57.070$ & $0.543$ & $0.461$& $61.227$ & $0.645$ & $0.632$\\
 			Ouni08\tnote{1}~\cite{ouni2008new}& CIELAB & $29.977$ & $0.826$ & $0.821$& $30.355$ & $0.668$ & $0.563$& $31.444$ & $0.726$ & $0.685$\\
 		    Simone09~\cite{simone2009alternative}  & OSA-UCS  & $35.798$ & $0.687$ & $0.697$& $35.212$ & $0.528$ & $0.395$& $36.712$ & $0.564$ & $0.545$\\
 			Pedersen12~\cite{pedersen2012new}&  CIELAB & $60.385$ & $0.798$ & $0.812$& $58.565$ & $0.482$ & $0.407$& $63.153$ & $0.612$ & $0.600$\\
 			Lee14~\cite{lee2014towards} & CIELAB  & $46.025$ & $0.571$ & $0.578$& $39.688$ & $0.282$ & $0.233$& $54.696$ & $0.350$ & $0.291$\\
 		    Jaramillo19~\cite{jaramillo2019evaluation} & YC$_{r}$C$_{b}$  &  $43.419$ & $0.514$ & $0.506$& $50.299$ & $0.081$ & $0.041$& $68.805$ & $0.321$ & $0.329$\\
 		    
 		    \hline
 		     SSIM~\cite{wang2004image} & Grayscale & $39.393$ & $0.589$ & $0.549$& $53.035$ & $0.077$ & $0.044$& $48.025$ & $0.309$ & $0.324$\\
 			Pinson04~\cite{pinson2004new} & YC$_{r}$C$_{b}$  & $51.719$ & $0.312$ & $0.290$& $59.114$ & $0.100$ & $0.084$& $59.021$ & $0.230$ & $0.207$\\
 			Yu09~\cite{yu2009method}& HSI & $69.289$ & $0.306$ & $0.319$& $67.511$ & $0.268$ & $0.233$& $68.891$ & $0.278$ & $0.297$\\
 		    Ponomarenko11~\cite{ponomarenko2011modified} & YC$_{r}$C$_{b}$ &  $50.315$ & $0.527$ & $0.536$& $47.826$ & $0.125$ & $0.104$& $52.687$ & $0.307$ & $0.293$\\
 			Gao13~\cite{gao2013no}& OCC & $63.498$ & $0.243$ & $0.207$& $60.698$ & $0.343$ & $0.246$& $62.932$ & $0.279$ & $0.226$\\
 		    VSI~\cite{zhang2014vsi} & LMN & $35.221$ & $0.617$ & $0.665$& $39.033$ & $0.160$ & $0.114$& $36.482$ & $0.404$ & $0.391$\\
 		    \FLIP~\cite{andersson2020flip} & CIELAB & $29.318$ & $0.745$ & $0.715$ & $27.158$ & $0.734$ & $0.640$& $29.099$ & $0.718$ & $0.663$\\
 		    PieAPP~\cite{prashnani2018pieapp} & RGB & $41.258$ & $0.510$ & $0.517$ & $38.457$ & $0.502$ & $0.433$& $41.375$ & $0.478$ & $0.460$\\
 		    LPIPS~\cite{zhang2018lpips} & RGB & $47.340$ & $0.674$ & $0.683$ & $40.104$ & $0.258$ & $0.239$& $66.594$ & $0.428$ & $0.439$\\   
 		    DISTS~\cite{ding2020image} & RGB & $39.771$ & $0.735$ & $0.730$ & $38.247$ & $0.428$ & $0.388$& $52.413$ & $0.437$ & $0.384$\\
 		   \hline
 			Chou07~\cite{chou2007fidelity}  & CIELAB  &  $50.721$ & $0.787$ & $0.785$& $36.184$ & $0.603$ & $0.459$& $49.545$ & $0.612$ & $0.557$\\
 			Lissner12~\cite{lissner2012image} &CIELAB&  $36.810$ & $0.605$ & $0.618$& $40.144$ & $0.339$ & $0.247$& $41.449$ & $0.429$ & $0.420$\\ 
 		    Butteraugli~\cite{alakuijala2017guetzli} & RGB & $42.620$ & $0.606$ & $0.593$ & $48.217$ & $0.258$ & $0.245$& $54.737$ & $0.371$ & $0.359$\\
 		   \hline
 		   VGG~\cite{simonyan2015vgg} & RGB & $19.199$&  $0.843$& $0.831$& $24.052$& $\textbf{0.833}$& $0.771$& $\textbf{20.906}$& $0.836$& $0.814$\\
 		   ResNet-18~\cite{he2016resnet} & RGB & $\textbf{17.969}$  & $\textbf{0.883}$ &$\textbf{0.892}$ &$\textbf{19.577}$ & $\textbf{0.874}$ & $\textbf{0.849}$& $\textbf{18.574}$ & $\textbf{0.876}$ & $\textbf{0.889}$\\
 		   UNet~\cite{ronneberger2015unet} & RGB & $\textbf{18.236}$& $0.849$& $0.843$& $26.039$& $0.789$& $0.760$ & $21.073$ &$0.813$ &$0.812$ \\
 		   CAN~\cite{chen2017fast} & RGB & $19.826$ & $0.858$ & $0.861$ & $23.158$ & $0.825$ & $0.744$ & $21.152$ &${0.833}$ & $0.818$\\
			\hline
			CD-Net & RGB &  $20.891$  & $\textbf{0.867}$ & $\textbf{0.870}$ & $\textbf{22.543}$ & $0.818$ & $\textbf{0.776}$& $21.431$ & $\textbf{0.846}$ & $\textbf{0.842}$\\  
			\bottomrule[1pt]
		\end{tabular}
		\begin{tablenotes}
            \item[1] The spatial extension of CIEDE2000.
        \end{tablenotes}

    \end{threeparttable}
	\end{center}
\end{table*}

\section{Experiments}
In this section, we first describe the training and testing procedures of the proposed CD-Net for CD assessment. We then quantitatively compare CD-Net with $33$ existing CD measures, and qualitatively examine the generated CD maps. We last conduct ablation studies to justify the key design choices of CD-Net, and probe its generalization on the COM dataset~\cite{luo2001ciede2000} and as a ``proper'' metric.
\subsection{CD-Net Training and Testing}
\label{sec:cd-net}
The training of CD-Net is carried out by minimizing the mean squared error (MSE) between its predictions and the ground-truth perceptual CDs over a mini-batch $\mathcal{B}$ sampled from SPCD:
\begin{align}
    \ell = \frac{1}{\vert\mathcal{B}\vert}\sum_{i=1}^{\vert\mathcal{B}\vert}\Vert \varDelta E^{(i)} - \varDelta V^{(i)}\Vert_2^2.
    \label{eq:loss}
\end{align}
 The Adam method is used as the stochastic optimizer with an initial learning rate of $10^{-3}$, a mini-batch size of $8$ and a decay factor of $2$ for every $50$ epochs, and we train CD-Net for $100$ epochs. We randomly sample $70\%$, $10\%$, and $20\%$ image pairs in SPCD as training, validation, and test sets, respectively, while ensuring content independence. During training, we crop the image to $768\times768$, and keep the original size for testing. In all experiments, we select the model with the best validation performance on images of the original size for testing. To reduce the bias caused by the randomness in training, validation and test set splitting, we repeat the whole procedure ten times, and report the mean results. 
 
Three criteria are used to quantitatively evaluate the performance of CD-Net: STRESS~\cite{garcia2007measurement}, SRCC, and PLCC. STRESS is proposed in~\cite{garcia2007measurement} as both prediction accuracy and statistical significance measure:
\begin{align}
            \mathrm{STRESS} = 100\sqrt{\frac{\sum_{i=1}^{M}(\varDelta E_i - F \varDelta V_i)^2}{F^2 \sum_{i=1}^{M}
        \varDelta V_i^2}},
\end{align}
        where $M$ is the number of test pairs and $F$ is the scale correction factor between $\varDelta E$ and $\varDelta V$, defined as
        \begin{align}
            F = \frac{\sum_{i=1}^{M}\varDelta E_i ^2}{\sum_{i=1}^{M} \varDelta E_i\varDelta V_i}.
        \end{align}
STRESS ranges from $0$ to $100$ with a small value indicating a tight fitting between model predictions and ground truths. SRCC and PLCC, on the other hand, measure prediction monotonicity and prediction linearity, respectively. PLCC is computed by 
        \begin{align}
        \label{eq:plcc}
            \mathrm{PLCC} = \frac{\sum_{i=1}^{M} (\varDelta E_i - \varDelta \overline{E})(\varDelta V_i - \varDelta \overline{V})}{\sqrt{\sum_{i=1}^{M} (\varDelta E_i - \varDelta \overline{E})^2} \sqrt{\sum_{i=1}^{M} (\varDelta V_i - \varDelta \overline{V})^2}},
        \end{align}
where $\varDelta \overline{E} = \frac{1}{M}\sum_{i=1}^{M} \varDelta E_i$ and $\varDelta \overline{V} = \frac{1}{M}\sum_{i=1}^{M} \varDelta V_i$, are the mean predicted and perceptual CDs, respectively. A pre-processing step is added to linearize model predictions by fitting a four-parameter monotonic function before computing PLCC
\begin{align}
    \varDelta \widehat{E}
    = (\eta_1 -\eta_2)/(1+\exp(-(\varDelta E-\eta_3)/|\eta_4|)) + \eta_2.
    \label{eq:nonlinear_mapping}
\end{align} 
where $\{\eta_i; i=1,2,3,4\}$ are the parameters to be fitted. SRCC is defined as 
\begin{align}
            \mathrm{SRCC} = 1 - \frac{6\sum_{i=1}^{M} d_i^2}{M(M^2-1)},
\end{align}
where $d_i$ is the difference between the $i$-th pair's rank orders in $\varDelta E$ and $\varDelta V$.

\begin{figure*}
    \centering
    \begin{minipage}[t]{1.08\textwidth}
        \centering
         \hskip-4em         \subfloat[\footnotesize{CIELAB~\cite{robertson1977cielab}}]{\includegraphics[width=0.22\textwidth]{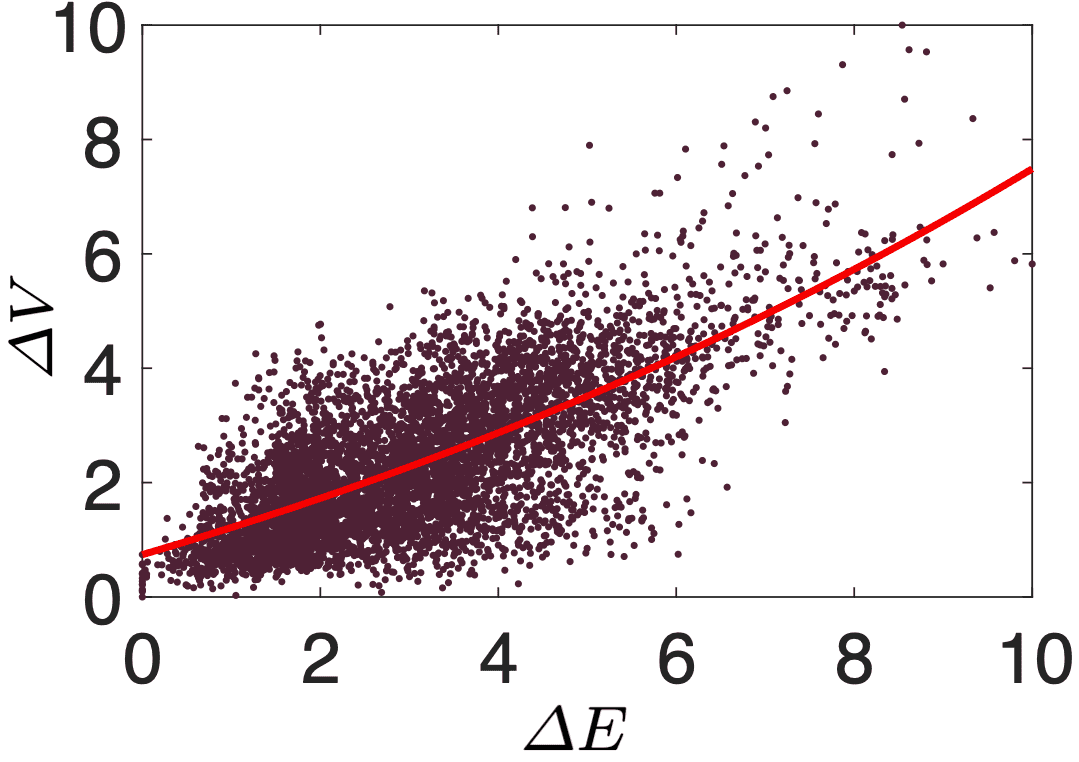}}\hskip.8em 
        \subfloat[\footnotesize{CIE94~\cite{mcdonald1995cie94}}]{\includegraphics[width=0.22\textwidth]{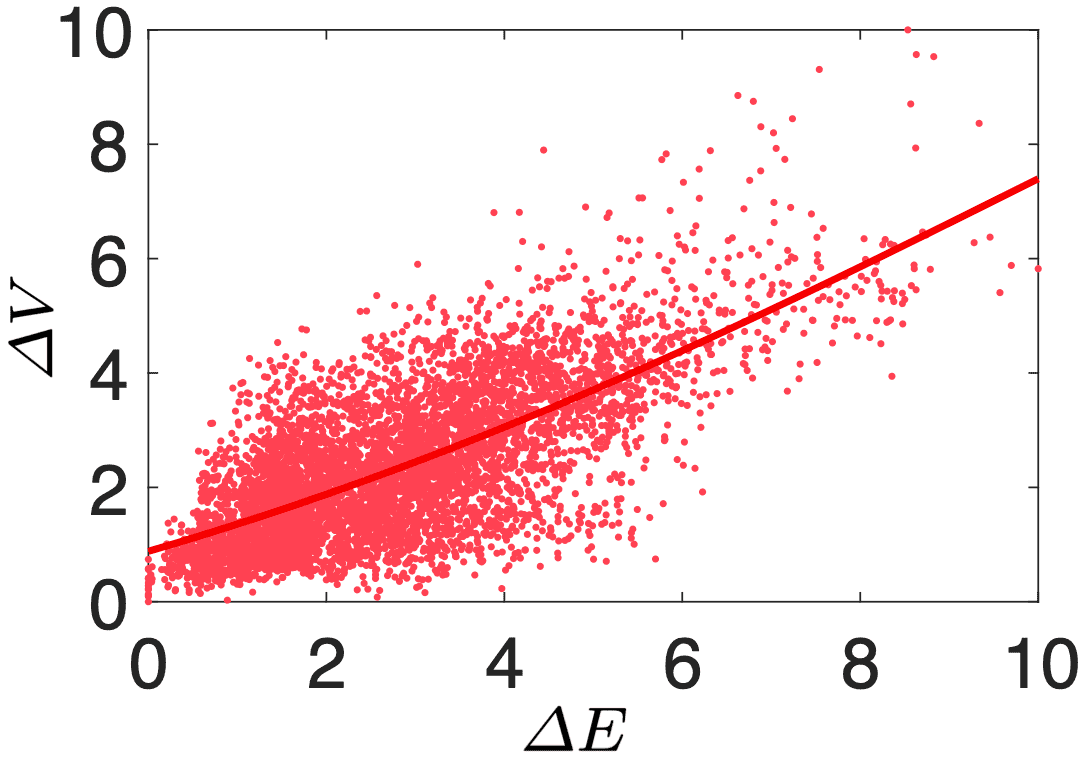}}\hskip.8em 
        \subfloat[\footnotesize{CIEDE2000~\cite{luo2001ciede2000}}]{\includegraphics[width=0.22\textwidth]{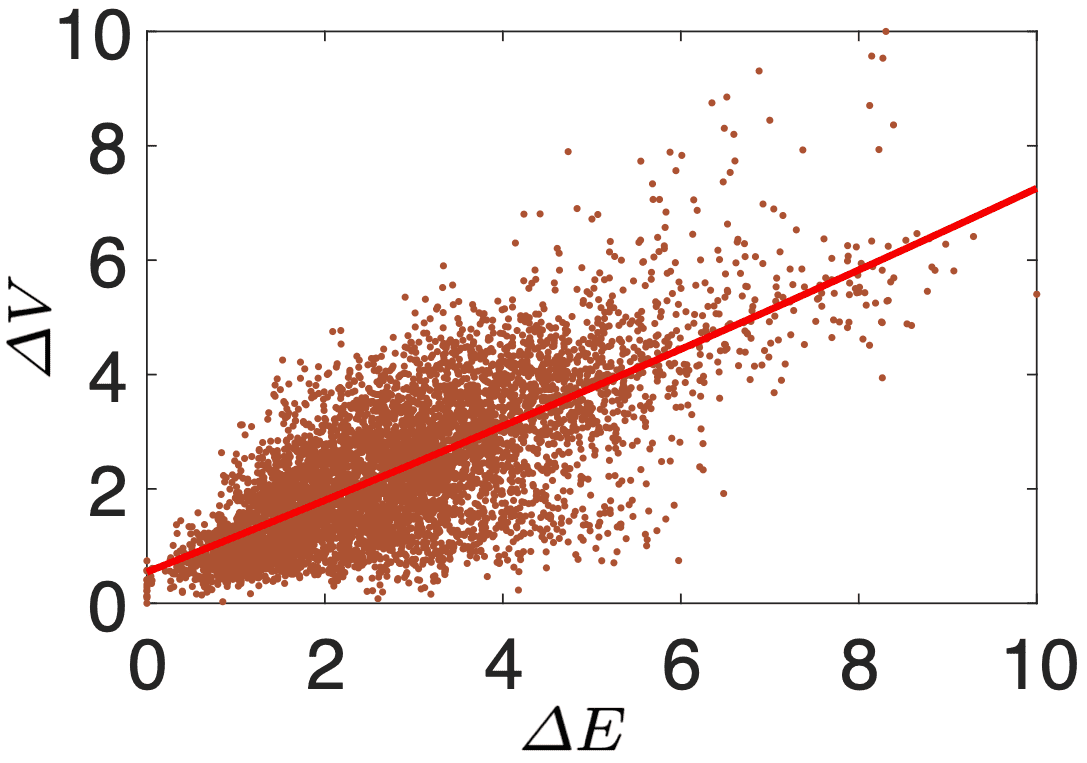}}\hskip.8em
        \subfloat[\footnotesize{Hong06~\cite{hong2006new}}]{\includegraphics[width=0.22\textwidth]{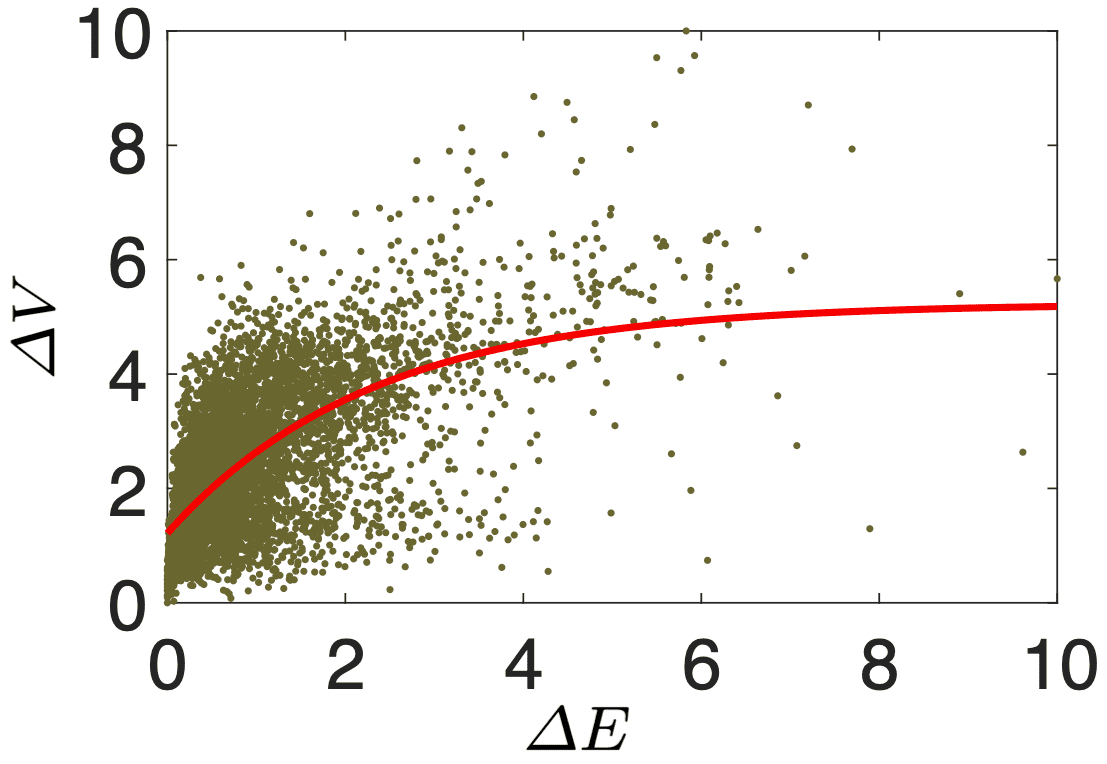}}\hskip.8em
    \end{minipage}
    \begin{minipage}[t]{1.08\textwidth}
        \centering
         \hskip-4em
         \subfloat[\footnotesize{\FLIP~\cite{andersson2020flip}}]{\includegraphics[width=0.22\textwidth]{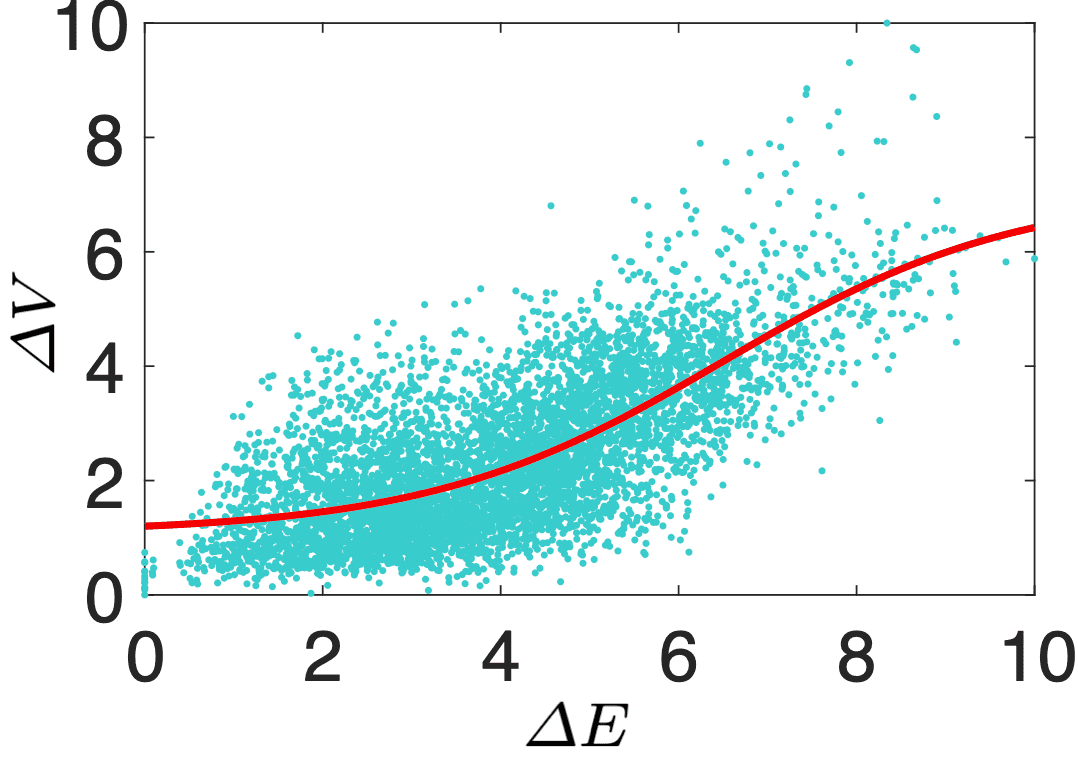}}\hskip.8em 
        \subfloat[\footnotesize{LPIPS~\cite{zhang2018lpips}}]{\includegraphics[width=0.22\textwidth]{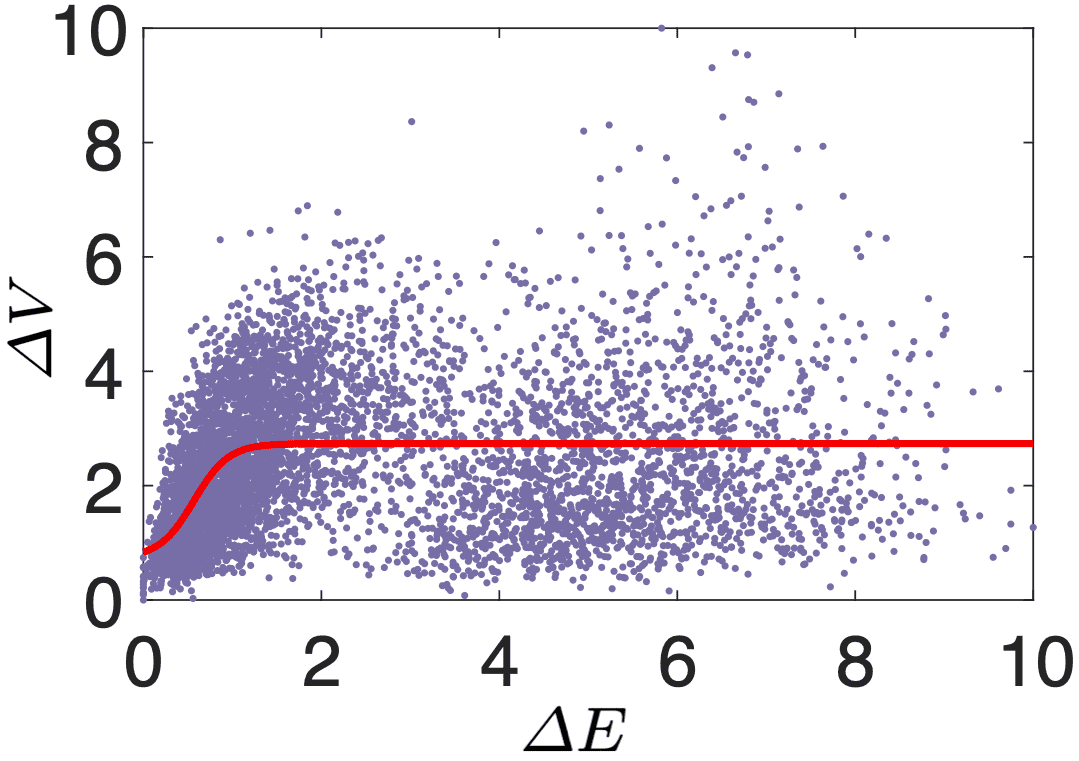}}\hskip.8em
        \subfloat[\footnotesize{DISTS~\cite{ding2020image}}]{\includegraphics[width=0.22\textwidth]{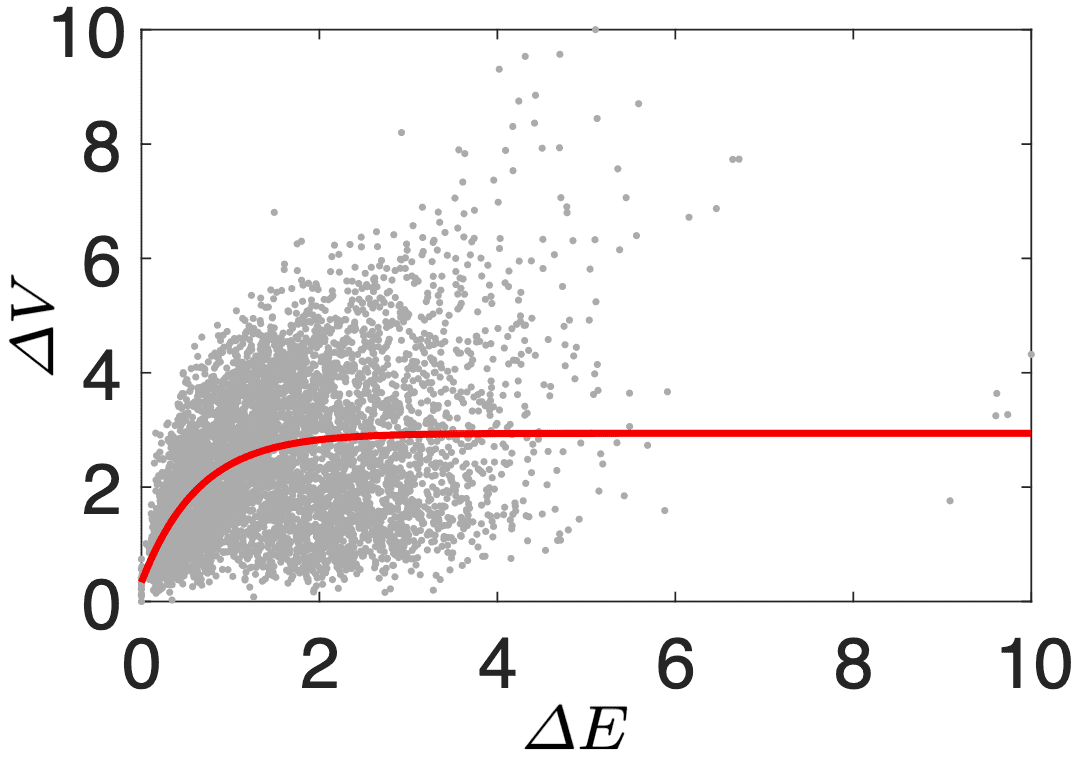}}\hskip.8em 
        \subfloat[\footnotesize{CD-Net}]{\includegraphics[width=0.22\textwidth]{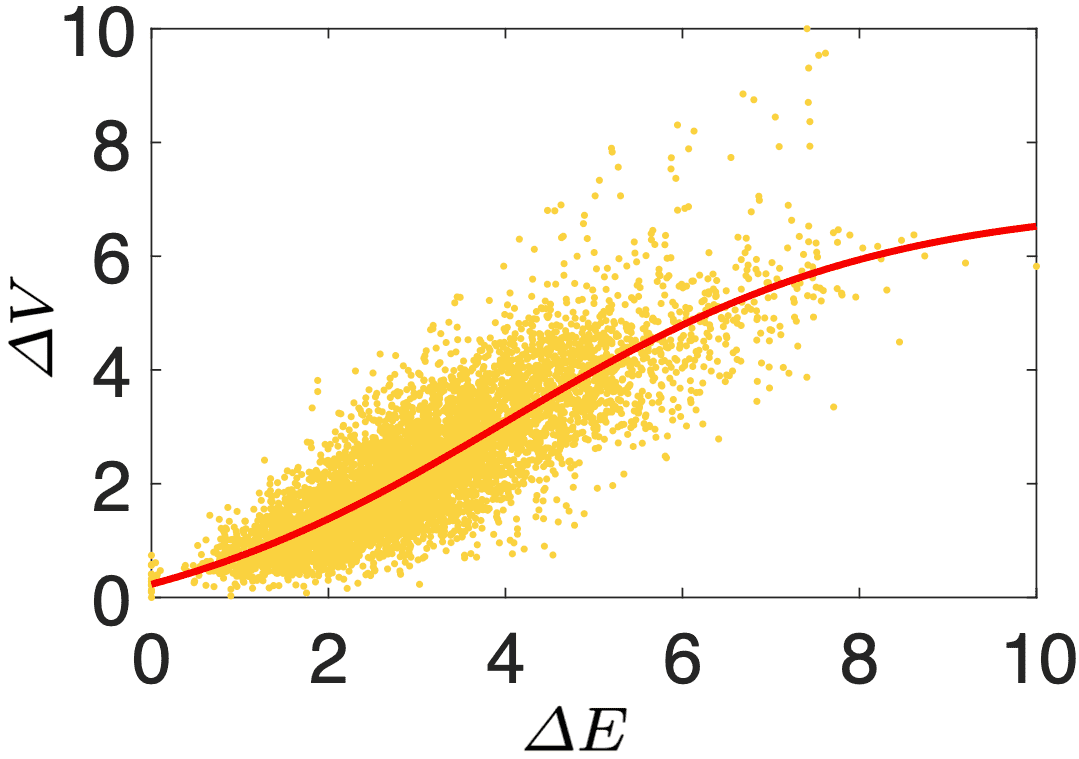}}\hskip.8em
    \end{minipage}
    \caption{Scatter plots of $\varDelta E$ against $\varDelta V$. } 
    \label{fig:scatter_plots}
\end{figure*}

\renewcommand{\thesubfigure}{\alph{subfigure}}
\begin{figure*}
    \centering
    \begin{minipage}[t]{1.08\textwidth}
        \subfloat[]{\includegraphics[width=0.165\textwidth]{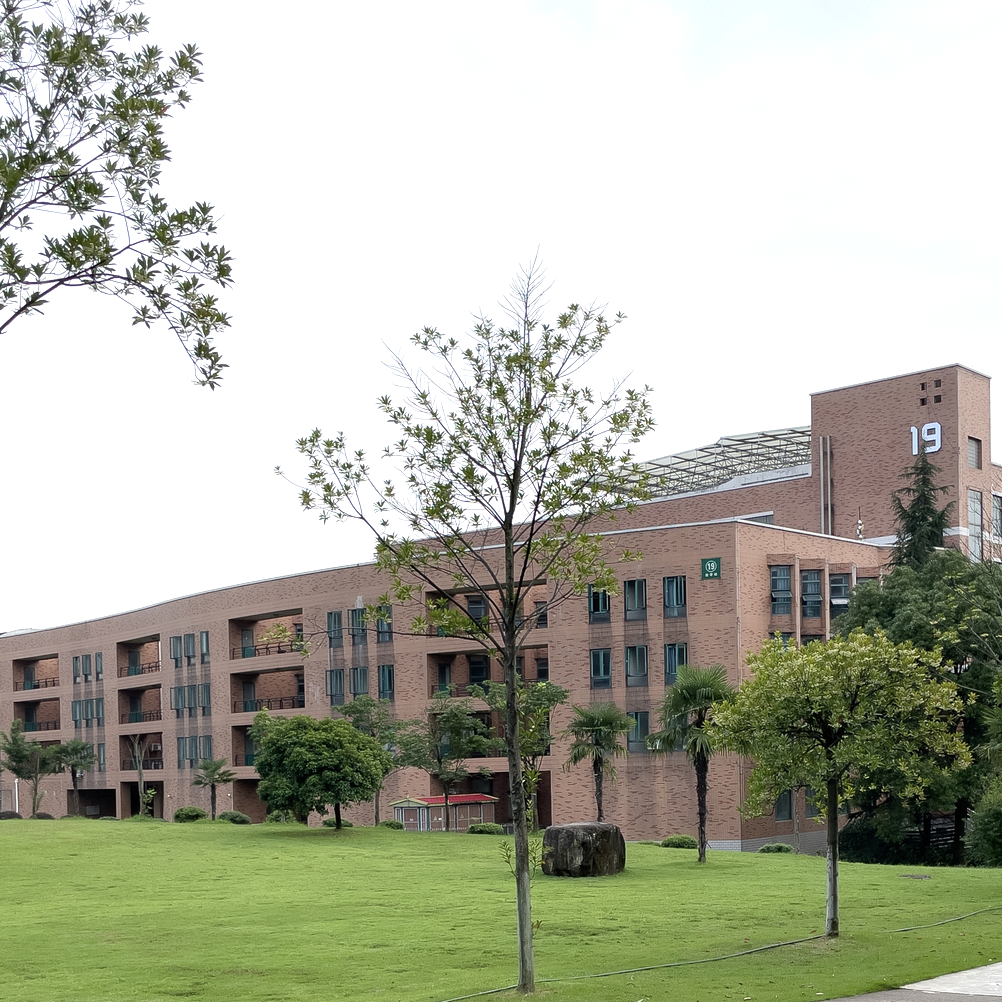}}\hskip.8em 
        \subfloat[\footnotesize{CIEDE2000~\cite{luo2001ciede2000}}]{\includegraphics[width=0.165\textwidth]{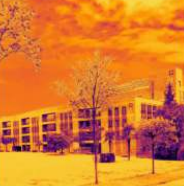}}\hskip.8em
        \subfloat[\footnotesize{S-CIELAB~\cite{zhang1997scielab}}]{\includegraphics[width=0.165\textwidth]{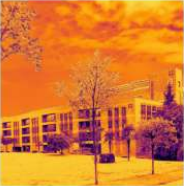}}\hskip.8em
        \subfloat[\footnotesize{Lee05~\cite{lee2005evaluation}}]{\includegraphics[width=0.165\textwidth]{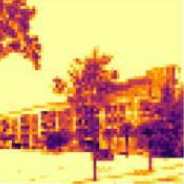}}\hskip.8em
        \subfloat[\footnotesize{\FLIP~\cite{andersson2020flip}}]{\includegraphics[width=0.165\textwidth]{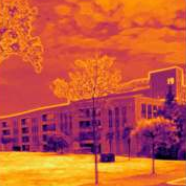}}\hskip.8em
    \end{minipage}
    \begin{minipage}[t]{1.08\textwidth}
        \subfloat[]{\includegraphics[width=0.165\textwidth]{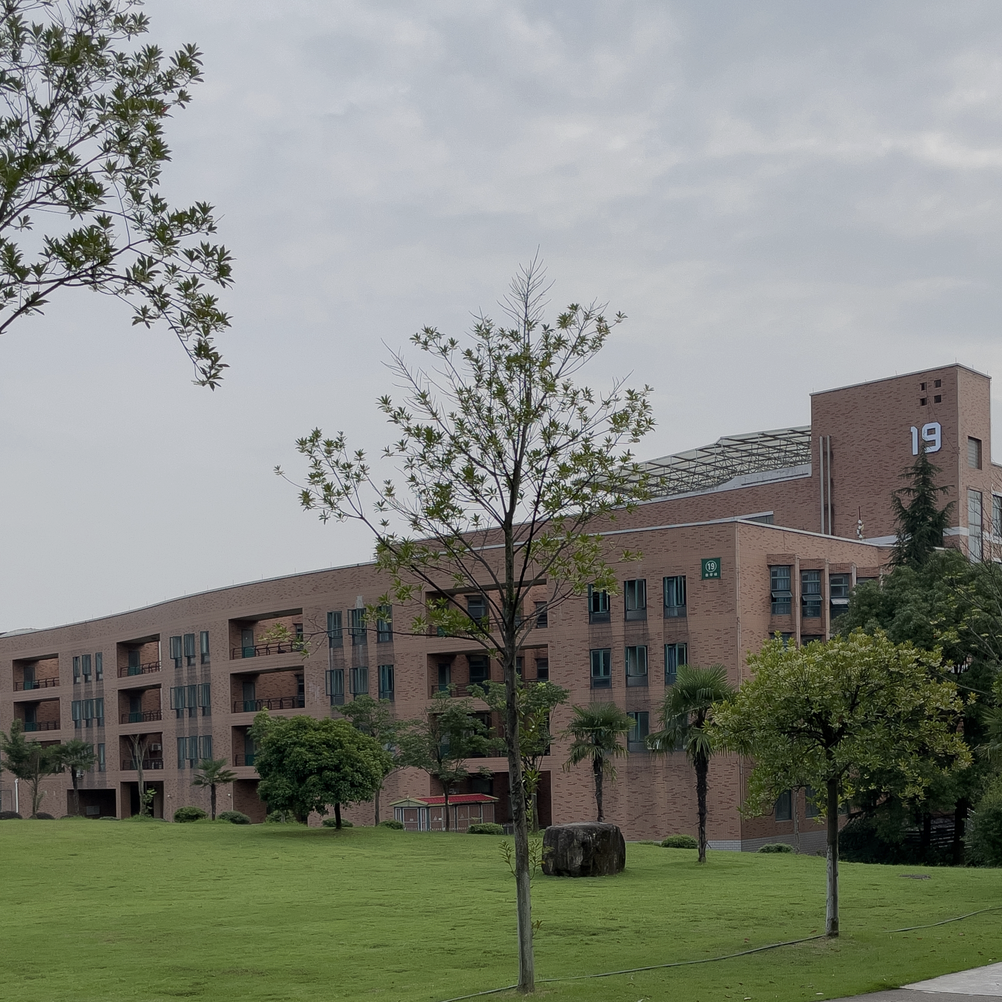}}\hskip.8em
        \subfloat[\footnotesize{ResNet-18~\cite{he2016resnet}}]{\includegraphics[width=0.165\textwidth]{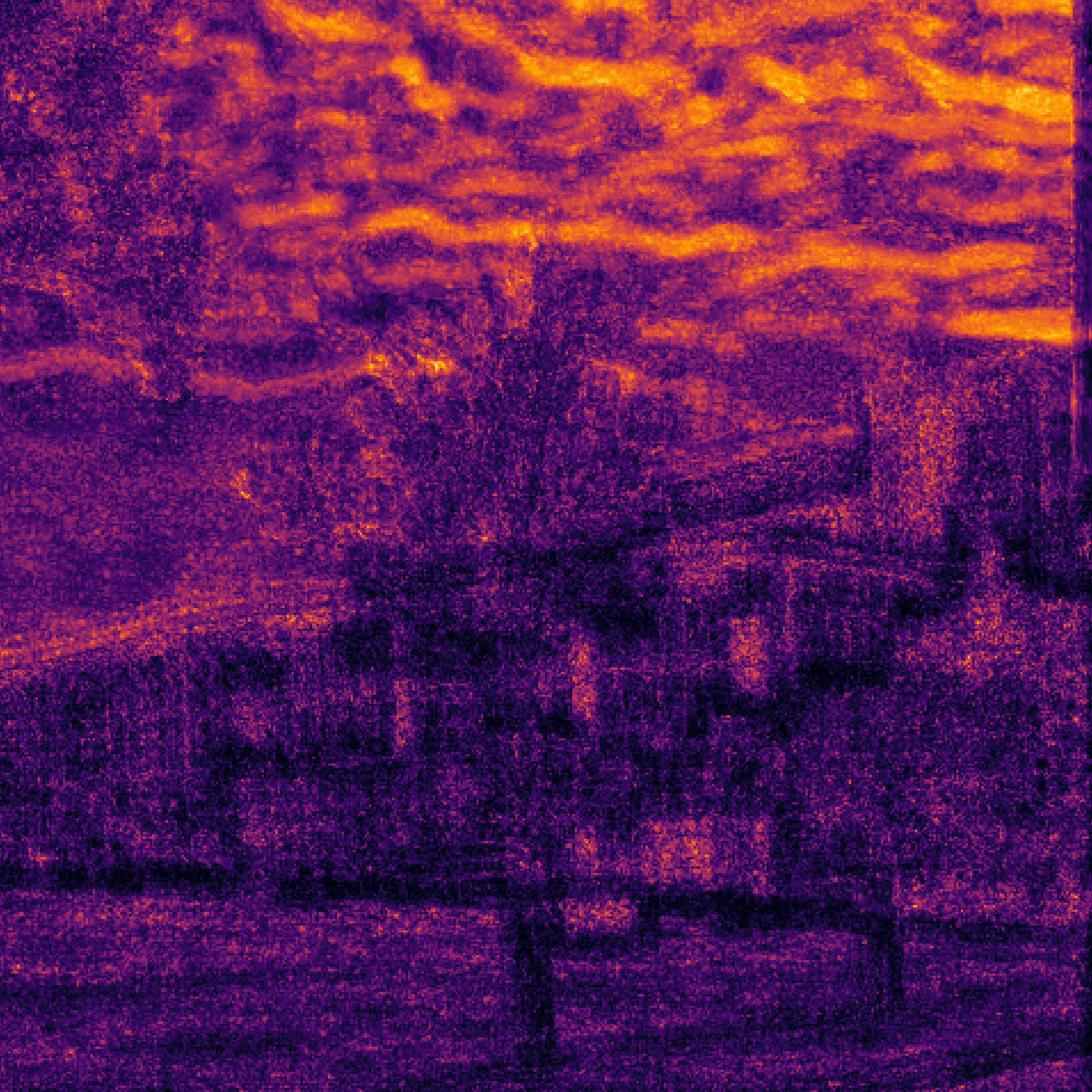}}\hskip.8em 
        \subfloat[\footnotesize{UNet~\cite{ronneberger2015unet}}]{\includegraphics[width=0.165\textwidth]{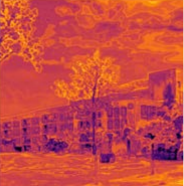}} \hskip.8em
        \subfloat[\footnotesize{CAN~\cite{chen2017fast}}]{\includegraphics[width=0.165\textwidth]{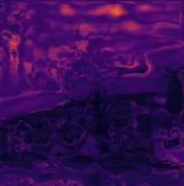}} \hskip.8em
        \subfloat[\footnotesize{CD-Net}]{\includegraphics[width=0.165\textwidth]{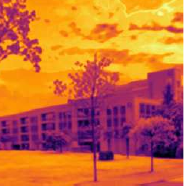}} \hskip.8em
        \subfloat[]{\includegraphics[width=0.0189\textwidth]{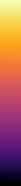}}
    \end{minipage}
    \caption{Visualization of local CD maps between two photographic images (a) and (f),  where a warmer color indicates a larger CD between two pixels/patches. For CD methods with different scales, we compensate it using Eq. \eqref{eq:nonlinear_mapping}, followed by linear scaling to the range of $[0, 255]$.}
    \label{fig:cd_maps} 
\end{figure*} 

\subsection{Main Results}
\noindent\textbf{Quantitative Results.} 
We compare CD-Net with $33$ existing CD measures, as listed in Table \ref{tab:comparision}. According to the initial developing objectives, we roughly categorize them into three classes, a) CD measures for homogeneous color patches and natural images - 1) CIELAB~\cite{robertson1977cielab},  2) CIE94~\cite{mcdonald1995cie94}, 3) CMC~\cite{bsi1998cmc}, 4) CIEDE2000~\cite{luo2001ciede2000}, 5) Huertas06~\cite{huertas2006performance}, 6) HyAB~\cite{abasi2020distance}, 7) HyCH~\cite{abasi2020distance}, 8) CIECAM02~\cite{luo2013ciecam02}, 9) CIECAM16~\cite{li2016ciecam16}, 10) J$_z$a$_z$b$_z$~\cite{safdar2017jzazbz}, 11) S-CIELAB~\cite{zhang1997scielab}, 12) Imai01~\cite{imai2001perceptual}, 13) Toet03~\cite{toet2003new}, 14) Lee05~\cite{lee2005evaluation}, 15) Hong06~\cite{hong2006new}, 16) Ouni08~\cite{ouni2008new}, 17) Simone09~\cite{simone2009alternative}, 18) Pedersen12~\cite{pedersen2012new}, 19) Lee14~\cite{lee2014towards} and 20) Jaramillo19~\cite{jaramillo2019evaluation}, b) general-purpose image quality models - 1) SSIM~\cite{wang2004image}, 2) Pinson04~\cite{pinson2004new}, 3) Yu09~\cite{yu2009method}, 4) Ponomarenko11~\cite{ponomarenko2011modified}, 5) Gao13~\cite{gao2013no}, 6) VSI~\cite{zhang2014vsi}, 7) \FLIP~\cite{andersson2020flip}, 8) PieAPP~\cite{prashnani2018pieapp}, 9) LPIPS~\cite{zhang2018lpips} and 10) DISTS~\cite{ding2020image}, and c) just noticeable difference (JND) measures - 1) Chou07~\cite{chou2007fidelity}, 2) Lissner12~\cite{lissner2012image} and 3) Butteraugli~\cite{alakuijala2017guetzli}. For HyAB, HyCH, CIECAM02, CIECAM16, J$_z$a$_z$b$_z$, S-CIELAB, Butteraugli, \FLIP, PieAPP, LPIPS, and DISTS, we use the official implementations provided by the respective authors, and for the rest methods, we employ the publicly available implementations\footnote{\url{https://telin.ugent.be/~bortiz/color_new}} provided by Jaramillo \etal~\cite{jaramillo2019evaluation}. We also train CNN models with frequently used backbones, \ie, VGG~\cite{simonyan2015vgg}, ResNet-18~\cite{he2016resnet}, UNet~\cite{ronneberger2015unet}, and CAN~\cite{chen2017fast}, as reference, whose trainable parameters are $14.715$ M, $11.177$ M, $31.038$ M and $37,600$, respectively. 

Table \ref{tab:comparision} reports the comparison results, from which we have several interesting observations. 
First, the CIE-recommended formulae (CIELAB~\cite{robertson1977cielab}, CIE94~\cite{mcdonald1995cie94}, CIEDE2000~\cite{luo2001ciede2000}, CIECAM02~\cite{luo2013ciecam02}, and CIECAM16~\cite{li2016ciecam16}), alongside their spatial extensions (S-CIELAB~\cite{zhang1997scielab} and Ouni08~\cite{ouni2008new}) perform favorably, which may benefit from the CIELAB space derived from solid color experiments for many years. Second, compared to CIEDE2000, the two spatial extensions, S-CIELAB and Ouni08, do not show noticeable improvements, and even degrade slightly in terms of STRESS, indicating that simple spatial filtering seems far from sufficient. Third, SSIM~\cite{wang2004image},     VSI~\cite{zhang2014vsi}, PieAPP~\cite{prashnani2018pieapp}, LPIPS~\cite{zhang2018lpips}, and DISTS~\cite{ding2020image}, which are tailored for image quality assessment, perform poorly when evaluating CDs. This is not surprising because the color alterations in SPCD may not be considered as visual distortions that lead to quality degradations. Fourth, JND measures that are designed to estimate the threshold at which a
difference is just barely visible fail to capture the suprathreshold CDs included in SPCD, leading to weak STRESS and correlation results. Fifth, the performance of most methods on perfectly aligned pairs is better than that on non-perfectly aligned ones, especially measured by correlation numbers, revealing their limitations in handling imperceptible geometric transformations. 
Finally, the proposed CD-Net is among the best performers, and even performs competitively against the four deep CNN-based models with a substantially larger number of parameters. 

Fig.~\ref{fig:scatter_plots} depicts the scatter plots between $\Delta E$ against $\Delta V$ of CIELAB~\cite{robertson1977cielab}, CIE94~\cite{mcdonald1995cie94}, CIEDE2000~\cite{luo2001ciede2000}, Hong06~\cite{hong2006new}, \FLIP~\cite{andersson2020flip}, LPIPS~\cite{zhang2018lpips}, DISTS~\cite{ding2020image}, and CD-Net. In these plots, each dot represents one test image pair, and the red curve is fitted by Eq. \eqref{eq:nonlinear_mapping}. It is clear that
CD-Net produces more linear and consistent predictions.
\\
\noindent\textbf{Qualitative Results.} 
Fig.~\ref{fig:cd_maps} illustrates the local CD maps, generated by eight representative CD formulae, \ie, CIEDE2000~\cite{luo2001ciede2000}, S-CIELAB~\cite{zhang1997scielab}, Lee05~\cite{lee2005evaluation}, \FLIP~\cite{andersson2020flip}, ResNet-18~\cite{chou2007fidelity}, UNet~\cite{ronneberger2015unet}, CAN~\cite{chen2017fast}, and the proposed CD-Net. To generate the CD map of ResNet-18, we first extract from an input image a set of $16\times 16$ overlapping patches with stride one (\ie, the $\mathrm{img2col}()$ operation in signal processing), and feed each patch to ResNet-18 for coordinate transform and local CD assessment, followed by $\mathrm{col2img}()$ operation with overlapped CD scores aggregated by simple averaging. We observe that CIEDE2000 generates a relatively noisy CD map due to pixel-wise comparison. The CD map generated by S-CIELAB, albeit with spatial filtering, does not exhibit much difference compared to CIEDE2000. The CD map of Lee05~\cite{lee2005evaluation} is of low-resolution, and suffers from blocky artifacts, which arises from $16 \times 16$ non-overlapping block processing. Besides, it fails to capture local CD variations (\eg, in the sky). \FLIP~\cite{andersson2020flip} tends to overemphasize CDs along salient edges (\eg, between the sky and the building).
Although ResNet-18, UNet, and CAN obtain similar or even better STRESS and correlation numbers compared to CD-Net, the generated CD maps are atypical. For example, the CD map by ResNet-18 looks noisy, while UNet generates ringing artifacts. This provides strong justifications that deep CNN-based methods are likely to overfit the training set, and may not learn the desired rules for CD assessment. In contrast, the proposed CD-Net generates reasonable CD maps, which well reflect the local CDs with smooth transitions across strong edges.  

\begin{table}[t]
\caption{Ablation analysis of the front-end filter bank}
    \label{tab:ablation_kernel}
    \vspace{-.3cm}
	\begin{center}
	    \begin{threeparttable} 
		\begin{tabular}{l|ccc}
    		\toprule[1pt]
			Convolution layer & STRESS$\downarrow$ & PLCC$\uparrow$ & SRCC$\uparrow$\\
           	\hline
 		    Only $1\times 1$ & $24.756$ & $0.785$ & $0.769$\\
 		    Only $11\times 11$ & $21.932$ & $0.838$ & $0.838$\\
 		   $1\times 1$ and $11\times 11$ & $21.431$ & $0.846$ & $0.842$\\
			\bottomrule[1pt]
		\end{tabular}
    \end{threeparttable}
	\end{center}
\end{table}

\begin{table}[t]
    \caption{Ablation analysis of the last convolution channel number}
    \label{tab:channel_comparision}
    \vspace{-.3cm}
	\begin{center}
	    \begin{threeparttable} 
		\begin{tabular}{c|ccc}
    		\toprule[1pt]
			\# of channels  & STRESS$\downarrow$ & PLCC$\uparrow$& SRCC$\uparrow$ \\
            \hline
            3 & $22.289$ & $0.843$ & $0.843$\\ 
            6 & $22.160$ & $0.842$ & $0.841$\\ 
            9 & $21.496$ & $0.844$ & $0.842$\\
            12 & $21.431$ & $0.846$ & $0.842$\\
            15 & $21.282$ & $0.845$ & $0.842$\\ 
            18 & $21.261$ & $0.845$ & $0.842$\\ 
			\bottomrule[1pt]
		\end{tabular}
    \end{threeparttable}
	\end{center}
\end{table}

\begin{table*}[t]
    \caption{Ablation analysis of the image size discrepancy during training and testing}
    \label{tab:ablation_study_image_size}
    \vspace{-.3cm}
    \begin{center}
        \hskip-.6cm
        \begin{tabular}{l|ccc|ccc|ccc|ccc}
            \toprule[1pt]
            \multirow{2}*{\diagbox{Train}{Test}} & \multicolumn{3}{c|}{256} & \multicolumn{3}{c|}{512} & \multicolumn{3}{c|}{768} & \multicolumn{3}{c}{1024}\\
            \cline{2-13}
             & STRESS & PLCC & SRCC & STRESS & PLCC & SRCC & STRESS & PLCC & SRCC & STRESS& PLCC & SRCC\\
            \hline
            256  & $21.789$ & $0.848$ & $0.850$& $21.904$ & $0.843$ & $0.843$& $22.107$ & $0.837$ & $0.836$& $22.291$ & $0.833$ & $0.829$\\
            512 & $22.306$ & $0.845$ & $0.850$& $22.011$ & $0.845$ & $0.848$& $22.013$ & $0.842$ & $0.843$& $22.072$ & $0.839$ & $0.837$\\
            768 & $21.643$ & $0.850$ & $0.852$& $21.346$ & $0.850$ & $0.852$& $21.376$ & $0.848$ & $0.847$& $21.431$ & $0.846$ & $0.842$\\
            1024 & $22.411$ & $0.848$ & $0.853$& $21.934$ & $0.849$ & $0.854$& $21.842$ & $0.848$ & $0.851$& $21.825$ & $0.845$ & $0.846$\\
            \bottomrule[1pt]
        \end{tabular}
    \end{center}
\end{table*}

\begin{table}[t]
    \caption{Generalizability evaluation on the TID2013 subset, which contains three types of color-related  distortions: quantization noise, image color quantization with dither, and chromatic aberration}
    \label{tab:comparision_tid2013}
    \vspace{-.3cm}
	\begin{center}
		\begin{tabular}{l|ccc}
    		\toprule[1pt]
			Method & STRESS$\downarrow$ & PLCC$\uparrow$ & SRCC$\uparrow$ \\
           \hline
 		CIELAB~\cite{robertson1977cielab}  & $18.795$ & $0.709$ & $0.714$  \\
 		CIE94~\cite{mcdonald1995cie94} & $18.673$ & $0.714$ & $0.718$\\
            CIEDE2000~\cite{luo2001ciede2000} & $19.198$ & $0.694$ & $0.703$\\
            HyCH~\cite{abasi2020distance} & $18.123$ & $0.733$ & $0.744$\\
            CIECAM02~\cite{luo2013ciecam02} & $18.336$ & $0.726$ & $0.732$ \\
            CIECAM16~\cite{li2016ciecam16} & $18.732$ & $0.711$ & $0.716$\\
            J$_z$a$_z$b$_z$~\cite{safdar2017jzazbz} & $20.380$ & $0.645$ & $0.654$\\
 		\hline
 	    S-CIELAB~\cite{zhang1997scielab} & $17.647$ & $0.749$ & $0.764$\\
 		SSIM~\cite{wang2004image}  & $17.022$ & $0.770$ & $0.776$\\
            VSI~\cite{zhang2014vsi}  & $15.967$ & $0.801$ & $0.780$\\ 
            \FLIP~\cite{andersson2020flip} & $12.610$ &$0.881$ & $0.882$ \\
 		PieAPP~\cite{prashnani2018pieapp}  & $20.316$ & $0.647$ & $0.634$\\
 		LPIPS~\cite{zhang2018lpips} & $14.157$ & $0.847$ & $0.842$\\
 		DISTS~\cite{ding2020image} & $15.645$ & $0.810$ & $0.796$\\
			\hline
			CD-Net & $15.474$ & $0.814$ & $0.813$\\  
			\bottomrule[1pt]
		\end{tabular}
	\end{center}
\end{table}

\begin{table*}[t]
    \caption{Generalizability evaluation on the COM dataset and its four sub-datasets: BFD-P, Leeds, Witt, and RIT-DuPont. The top two methods targeted for natural photographic images are highlighted in boldface. PLCC on RIT-DuPont is not computable, thus indicated by ``---''}
    \label{tab:color_patch_comparision}
    \vspace{-.3cm}
	\begin{center}
	    \begin{threeparttable} 
		\begin{tabular}{l|cccccccc|cc}
    		\toprule[1pt]
    		\multirow{2}*{Method} & \multicolumn{2}{c}{BFD-P~\cite{luo1986bfdp}} & \multicolumn{2}{c}{Leeds~\cite{kim1997leeds}} & \multicolumn{2}{c}{Witt~\cite{witt1999witt}} & \multicolumn{2}{c|}{RIT-DuPont~\cite{berns1991rit-dupont}} & \multicolumn{2}{c}{COM dataset~\cite{luo2001ciede2000}}\\
    		\cline{2-11} 
			 & STRESS$\downarrow$ & PLCC$\uparrow$ & STRESS$\downarrow$ & PLCC$\uparrow$ & STRESS$\downarrow$ & PLCC$\uparrow$ & STRESS$\downarrow$ & PLCC$\uparrow$ & STRESS$\downarrow$ & PLCC$\uparrow$\\
           	\hline
			CIELAB~\cite{robertson1977cielab} & $45.054$ & $0.749$ & $40.093$ & $0.295$ & $51.689$ & $0.565$ & $30.348$  & --- & $45.202$ & $0.693$ \\
			CIE94~\cite{mcdonald1995cie94} & $35.798$  & $0.830$ & $30.494$  & $0.584$ & $31.857$  & $0.793$ & $20.982$  & --- & $33.235$& $0.814$  \\
			CMC~\cite{bsi1998cmc} & $32.860$  & $0.852$ & $24.901$  & $0.698$ & $35.115$  & $0.758$ & $28.143$  & --- &$31.819$ & $0.831$ \\
            CIEDE2000~\cite{luo2001ciede2000} & $31.935$  & $0.861$ & $19.247$  & $0.772$ & $30.358$  & $0.825$ & $20.239$ & --- & $28.979$ & $0.862$ \\
 			\hline
 			Lee14~\cite{lee2014towards} & $98.316$ & $0.179$& $90.437$& $0.364$& $89.106$ & $0.516$& $96.390$& --- &$95.381$&$0.146$\\
 			Yu09~\cite{yu2009method}& $93.933$ & $0.375$ & $79.154$ & $0.381$ & $87.547$ & $0.479$ & $95.180$ & --- &$93.034$& $0.296$\\
 		    PieAPP~\cite{prashnani2018pieapp} & $83.719$ & $0.463$ & $74.009$ & $0.042$ & $74.837$ & $0.450$ & $77.024$ & --- & $81.835$ & $0.401$\\
 		    LPIPS~\cite{zhang2018lpips} & $81.684$ & $0.626$ & $ 75.784$ & $0.354$ & $79.894$ & $0.623$ & $ 86.057$ & --- & $81.913$ &$0.540$\\
 		    DISTS~\cite{ding2020image} & $66.756$ & $0.599$ &  $64.736$ & $\textbf{0.464}$ &  $66.396$ & $0.560$ & $76.368$ & --- &  $67.863$ & $0.536$ \\
 			Lissner12~\cite{lissner2012image} & $85.692$ & $0.297$ & $79.018$ & $0.267$ & $83.371$ & $0.397$ & $78.256$ & --- &$84.340$&$0.327$\\ 
     	    Butteraugli~\cite{alakuijala2017guetzli} & $52.711$ & $0.675$ & $44.924$ & $0.431$ & $50.767$ & $0.532$ & $36.370$ & --- & $50.161$& $0.611$\\
     	    \FLIP~\cite{andersson2020flip} &  $43.306$ & $0.693$ & $34.270$ & $0.310$ & $\textbf{37.310}$ & $0.705$ & $29.476$ & --- & $40.795$& $0.668$\\
 			VGG~\cite{simonyan2015vgg} & $44.542$ & $0.734$ & $\textbf{27.508}$ & $0.449$ &  $37.948$ & $0.728$ & $\textbf{17.166}$ & --- & $40.828$ & $0.691$\\
 		    ResNet-18~\cite{he2016resnet} & $\textbf{41.421}$ & $\textbf{0.778}$ & $\textbf{22.999}$ & $\textbf{0.568}$ & $38.224$ & $\textbf{0.811}$ &$\textbf{9.226}$ & --- &$\textbf{40.526}$ & $\textbf{0.775}$\\
 		    UNet~\cite{ronneberger2015unet} & $49.010$ & $0.611$ & $51.578$ & $0.289$& $56.436$ & $0.547$ & $54.523$ & --- & $57.775$ &$0.438$\\
 		    CAN~\cite{chen2017fast} & $51.305$ & $0.648$ & $48.988$ & $0.313$ & $41.845$ & $0.773$ & $51.632$ & --- & $49.407$& $0.640$ \\
 			\hline
			CD-Net & $\textbf{39.312}$ & $\textbf{0.791}$& $38.558$ & $0.449$ & $\textbf{33.640}$ & $\textbf{0.828}$ & $42.999$ & --- & $\textbf{38.872}$& $\textbf{0.786}$ \\
			\bottomrule[1pt]
		\end{tabular}
    \end{threeparttable}
	\end{center} 
\end{table*}

\subsection{Ablation Studies} 
In this subsection, we conduct ablation experiments to investigate the vital design choices of CD-Net, including the front-end filter bank, the last convolution channel number, and the training/testing image size discrepancy.
\vspace{0.4em} \\
\noindent \textbf{Front-end Filter Bank.} We compare different design choices for coordinate transform: using $1\times 1$ convolution filter only, using $11\times11$ convolution filter only, and concatenating the two convolution filters as the default setting. For a fair comparison, we adjust the output channel to be the same. Table \ref{tab:ablation_kernel} lists the results, where we find that using $1\times 1$ convolution layer only, similar to the CIE-recommended formulae, performs the worst, which is likely because humans do not perceive CDs in a pixel-wise fashion. Using an $11\times11$ convolution filter only improves the performance, which gives a better treatment of imperceptible geometric transformations. The default setting that concatenates the two filters performs the best, indicating that the two-scale computations may provide complementary information for CD assessment.
\vspace{0.4em} \\
\noindent \textbf{Last Convolution Channel Number.} We evaluate the influence of the channel number $C$ of the last convolution, \ie, the dimension of the feature space with possibly improved perceptual uniformity. Table \ref{tab:channel_comparision} reports the performance changes with varying $C$ sampled from \{$3$, $6$, $9$, $12$, $15$, $18$\}. We observe that CD-Net is fairly stable with respect to this hyperparameter.
\vspace{0.4em} \\
\noindent \textbf{Image Size. } We probe the effect of the image size discrepancy during training and testing on CD assessment. Specifically, we train and test CD-Net on images of different resolutions, and show the results in Table \ref{tab:ablation_study_image_size}. We find that CD-Net performance degrades slightly with the increase of testing image size when fixing the training image size. This may be because the problem of misalignment is more pronounced when the two images are of high resolution, leading to performance degradation on non-perfectly aligned pairs in SPCD. A straightforward remedy is to incorporate convolution filters of large kernel sizes into the front-end filter bank.
Besides, the performance of CD-Net is approximately proportional to the training image size, indicating that it is necessary to train models on images of high resolution (close to that used in subjective experiments).

\begin{figure*}
\hspace{1pt}
\begin{minipage}[c]{0.27\textwidth}
    \vspace{10pt}
    \subfloat[\footnotesize{Reference}]{\includegraphics[width=1.0\linewidth]{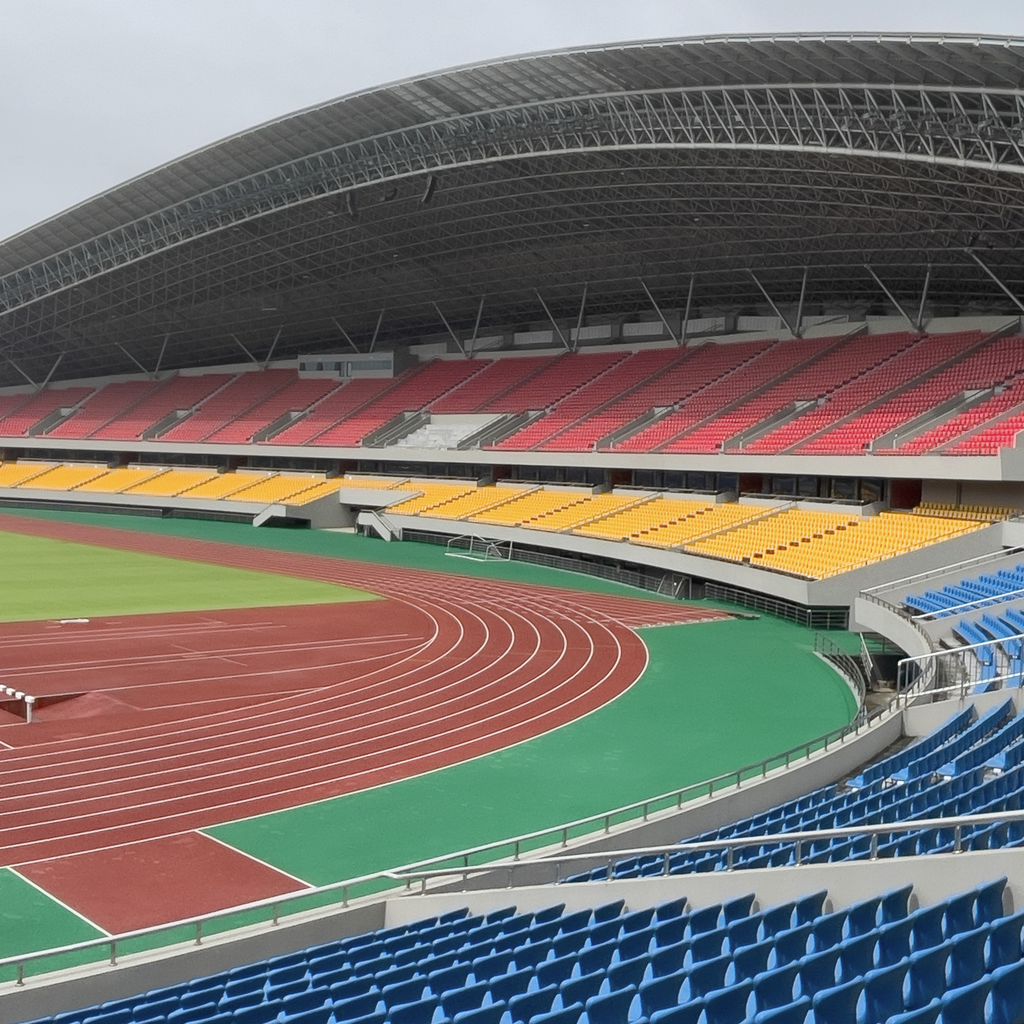}}
\end{minipage}
\hspace{0.5pt}
\begin{minipage}[c]{0.795\textwidth}
    \subfloat[\footnotesize{Noise}]{\includegraphics[width=0.14\linewidth]{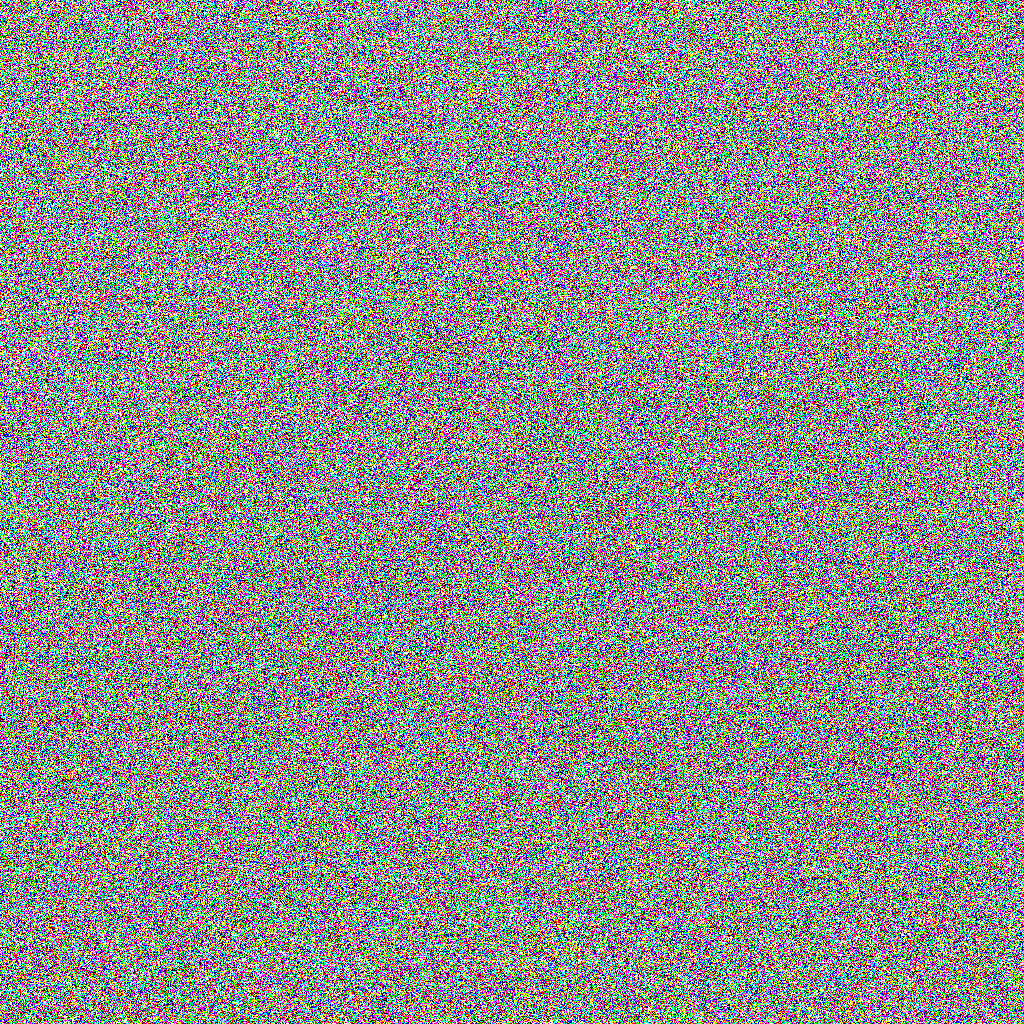}}
    \hspace{1pt}
    \subfloat[\footnotesize{VGG}]{\includegraphics[width=0.14\linewidth]{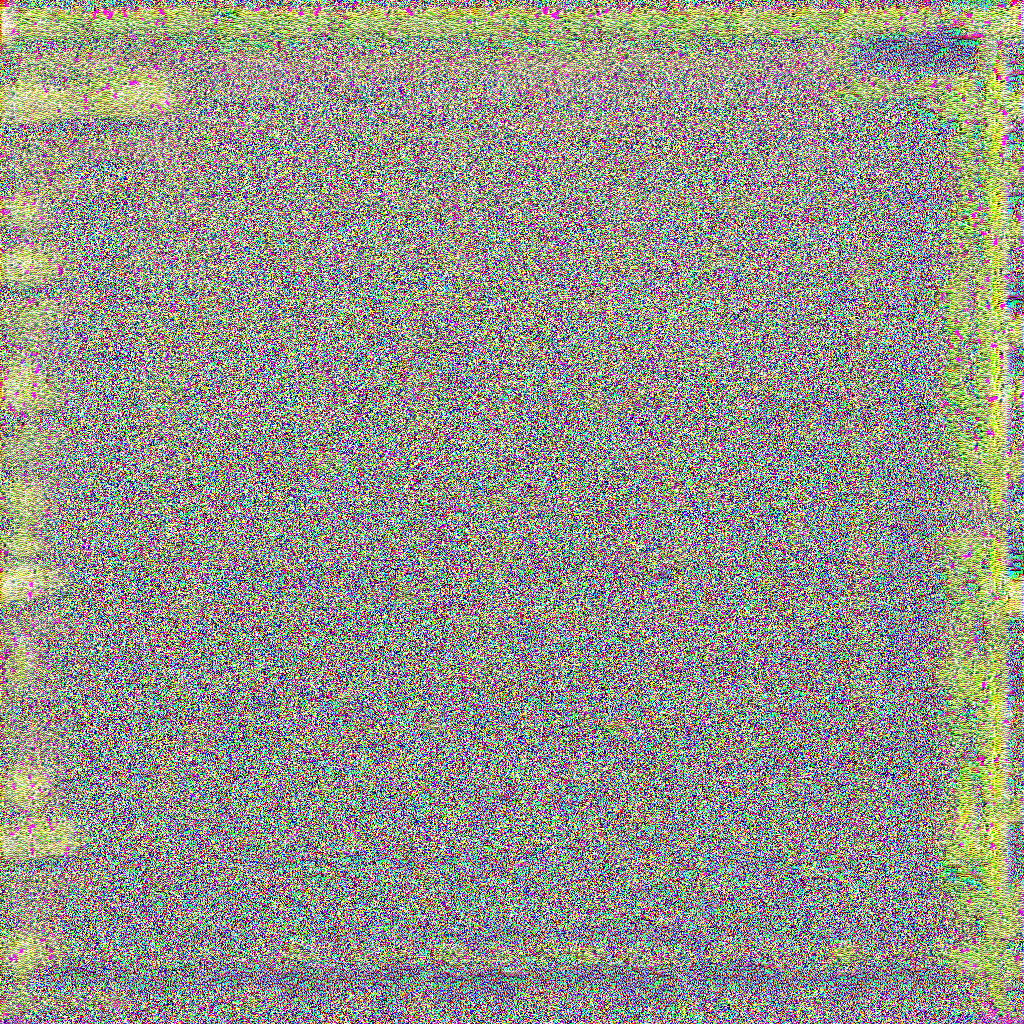}}
    \hspace{1pt}
    \subfloat[\footnotesize{ResNet-18}]{\includegraphics[width=0.14\linewidth]{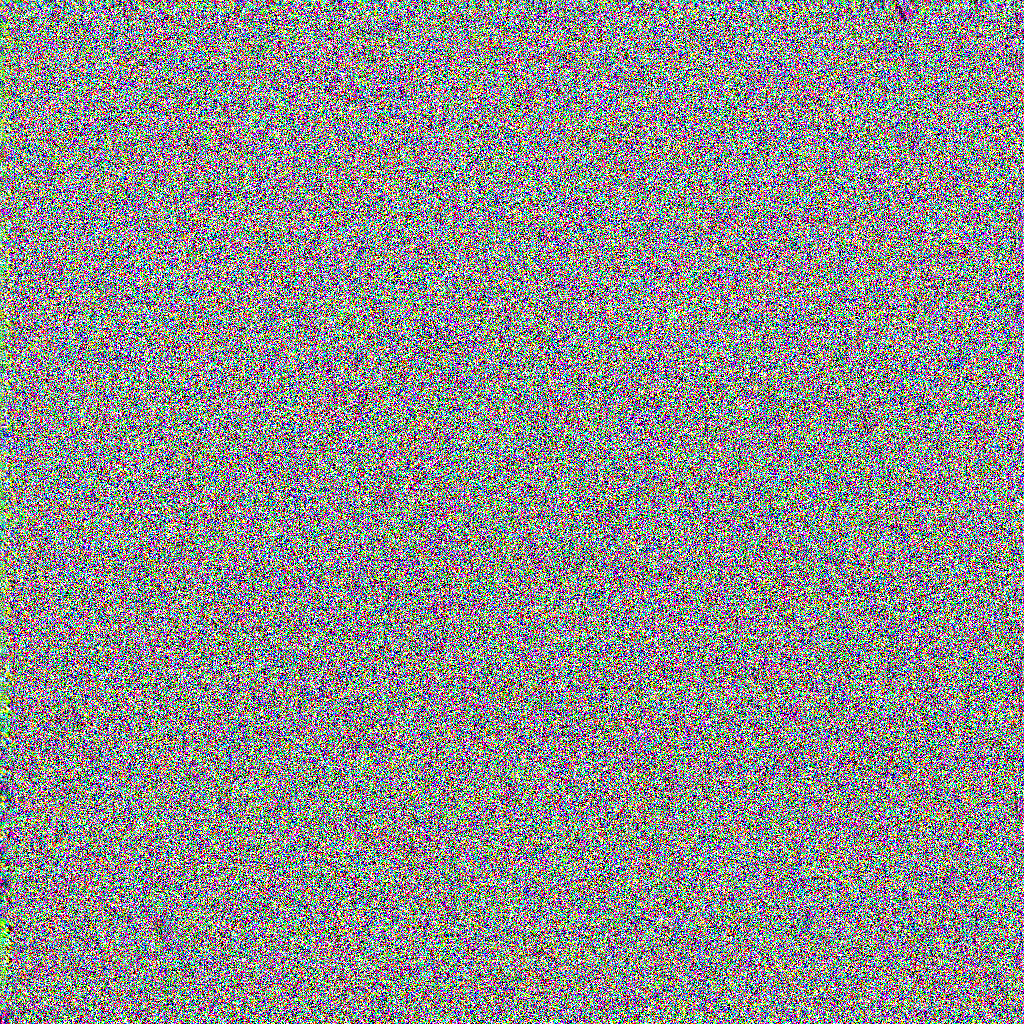}}
    \hspace{1pt}
    \subfloat[\footnotesize{UNet}]{\includegraphics[width=0.14\linewidth]{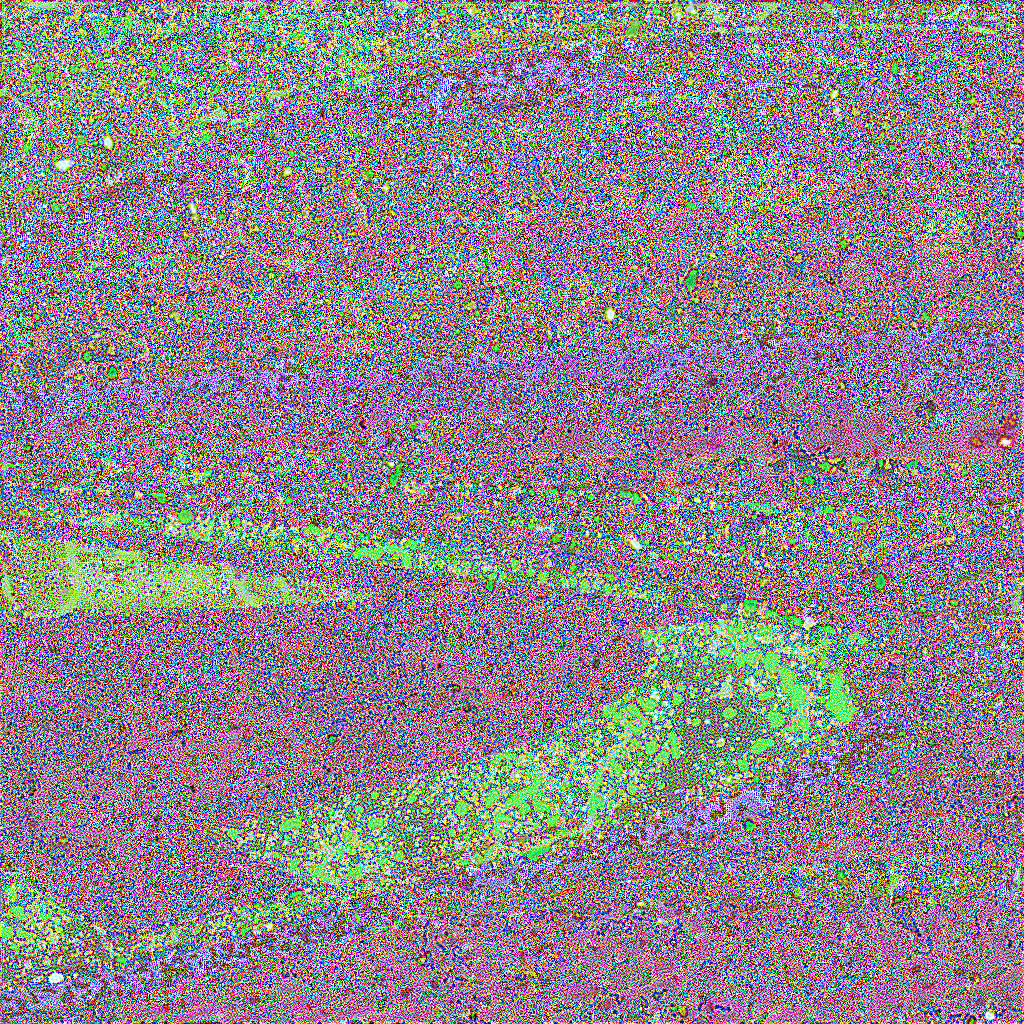}}
    \hspace{1pt}
    \subfloat[\footnotesize{CAN}]{\includegraphics[width=0.14\linewidth]{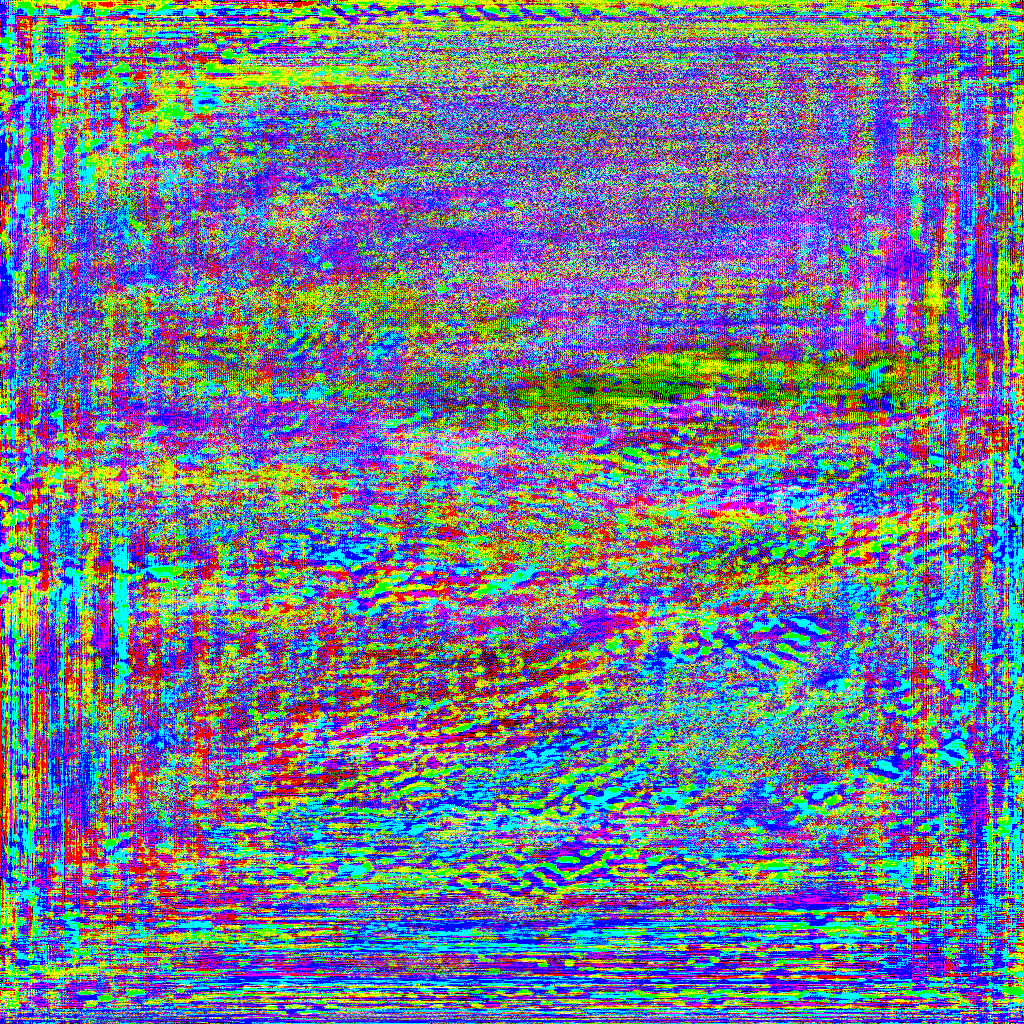}}
    \hspace{1pt}
    \subfloat[\footnotesize{CD-Net}]{\includegraphics[width=0.14\linewidth]{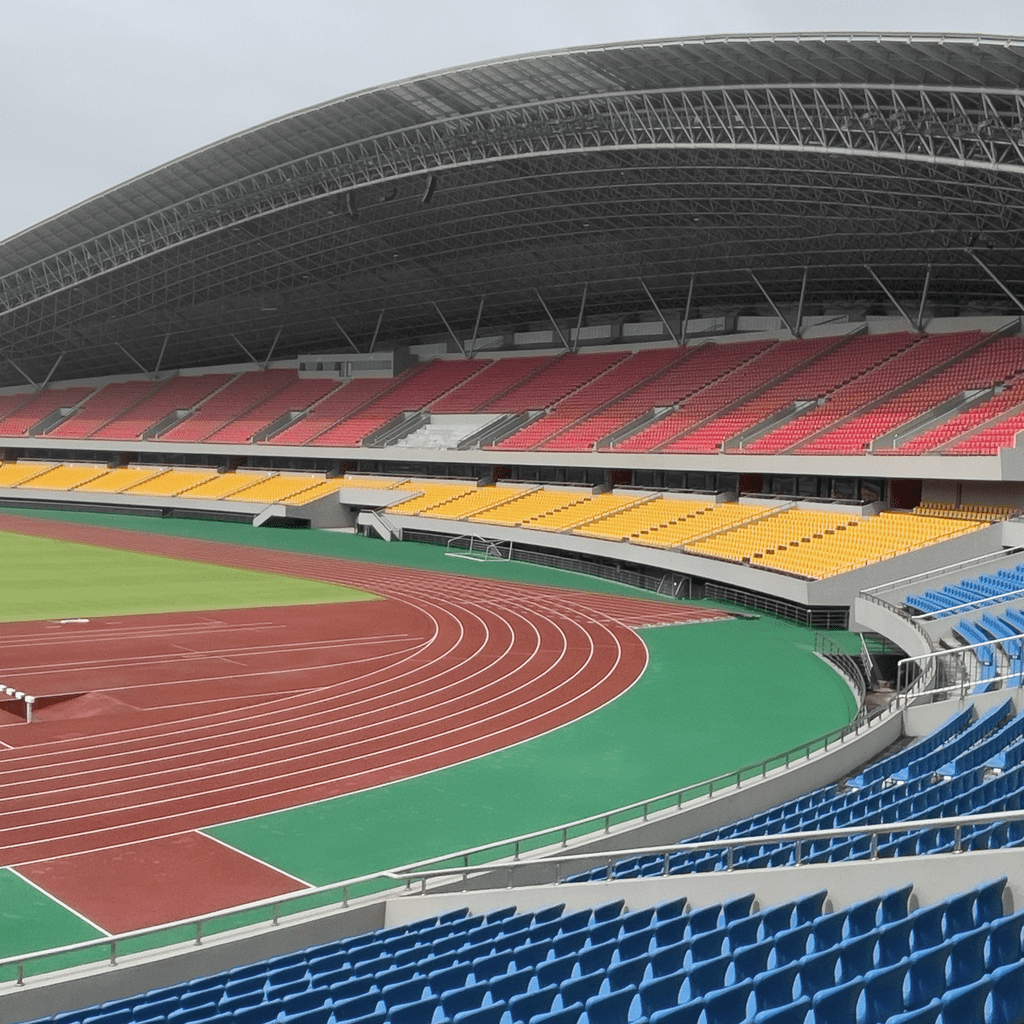}} \\
    \hspace{1pt}
   \subfloat[\footnotesize{Alternation}]{\includegraphics[width=0.14\linewidth]{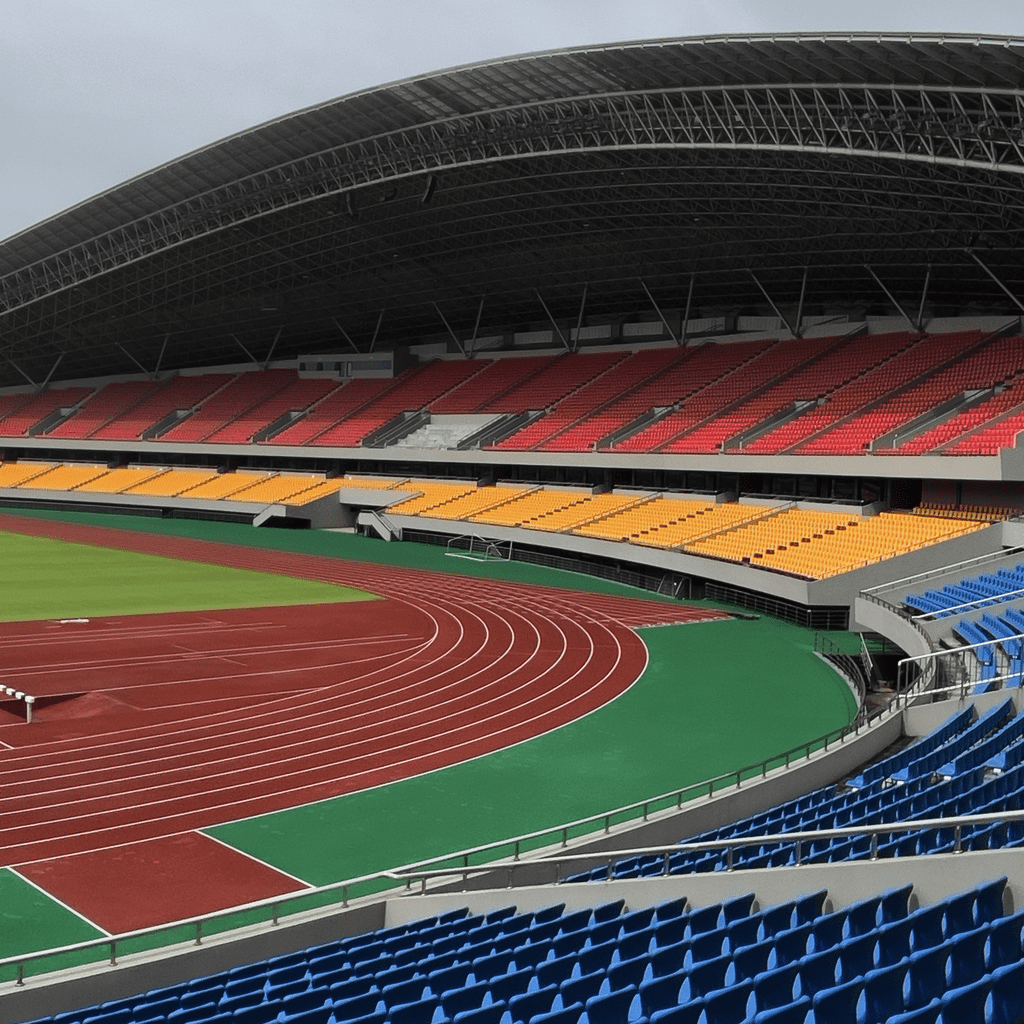}}
    \hspace{1pt}
    \subfloat[\footnotesize{VGG}]{\includegraphics[width=0.14\linewidth]{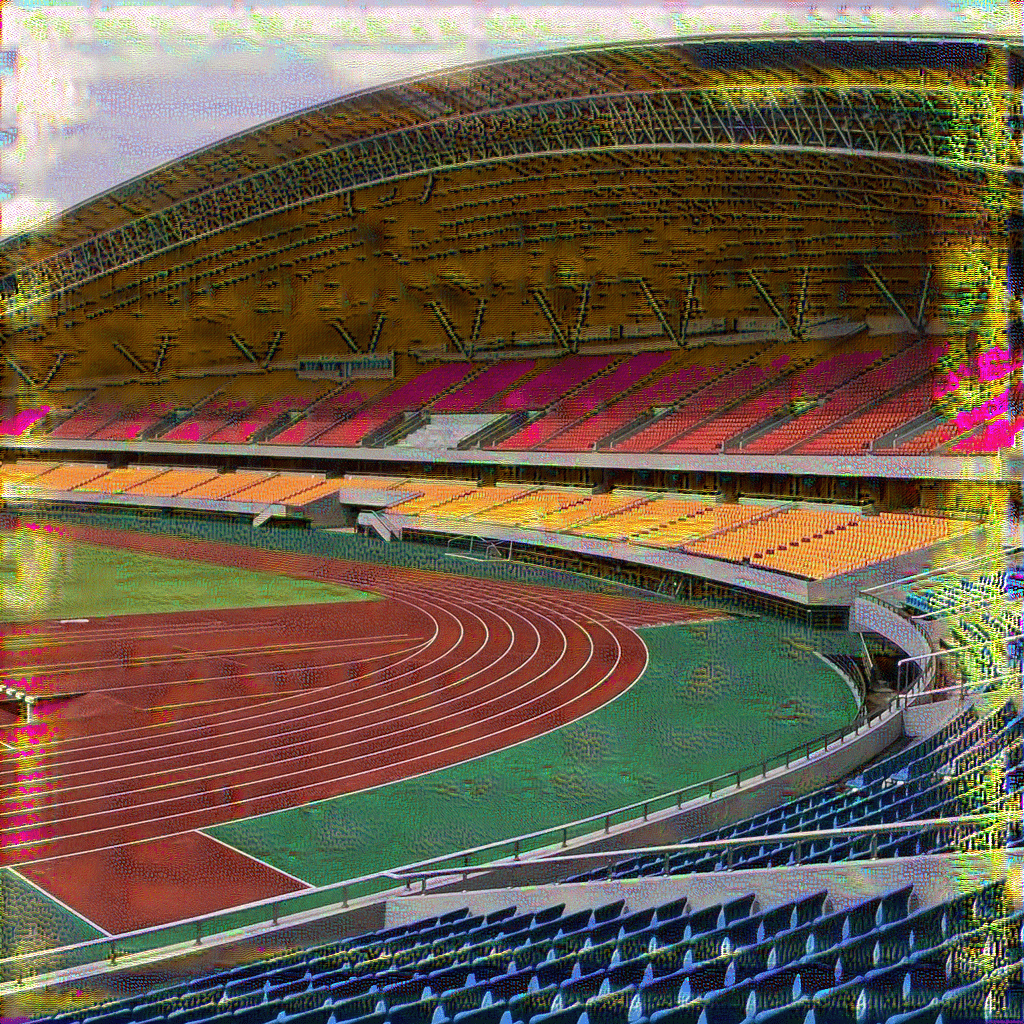}}
    \hspace{1pt}
    \subfloat[\footnotesize{ResNet-18}]{\includegraphics[width=0.14\linewidth]{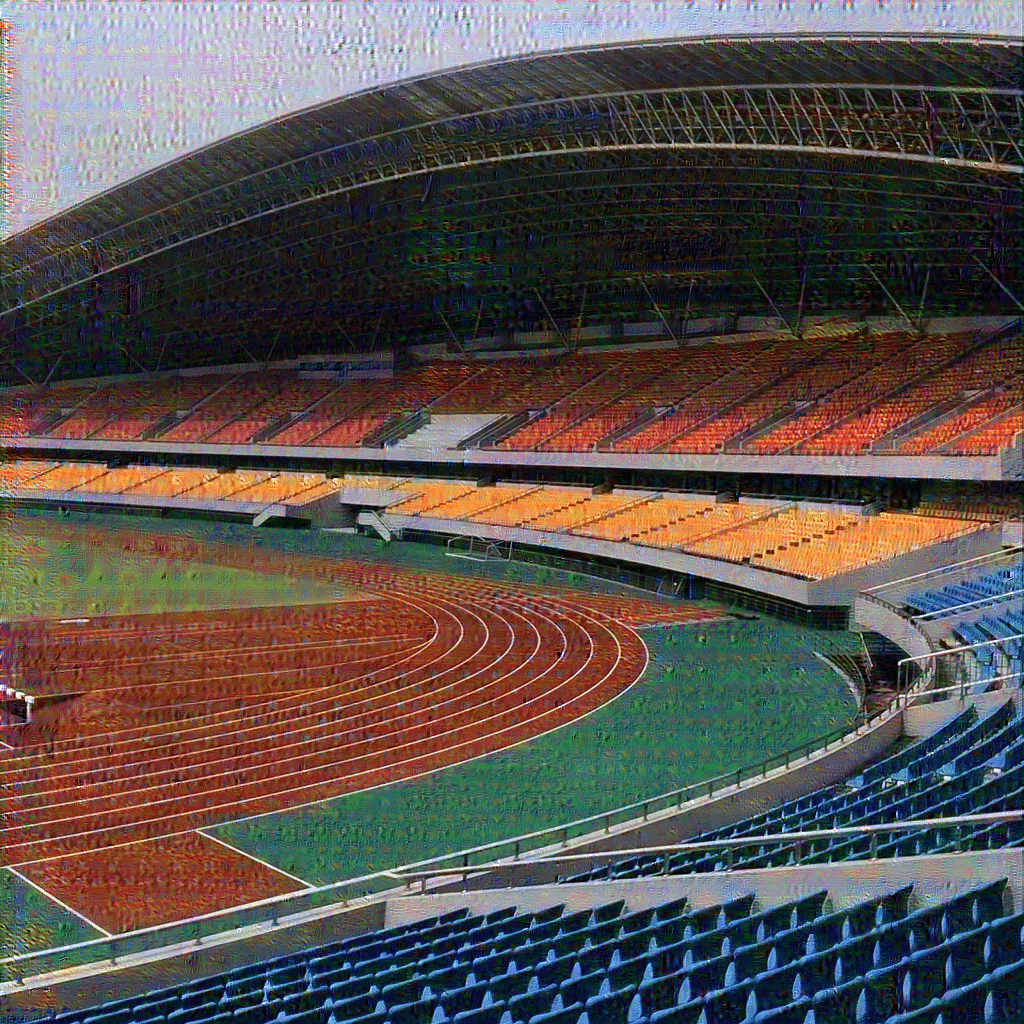}}
    \hspace{1pt}
    \subfloat[\footnotesize{UNet}]{\includegraphics[width=0.14\linewidth]{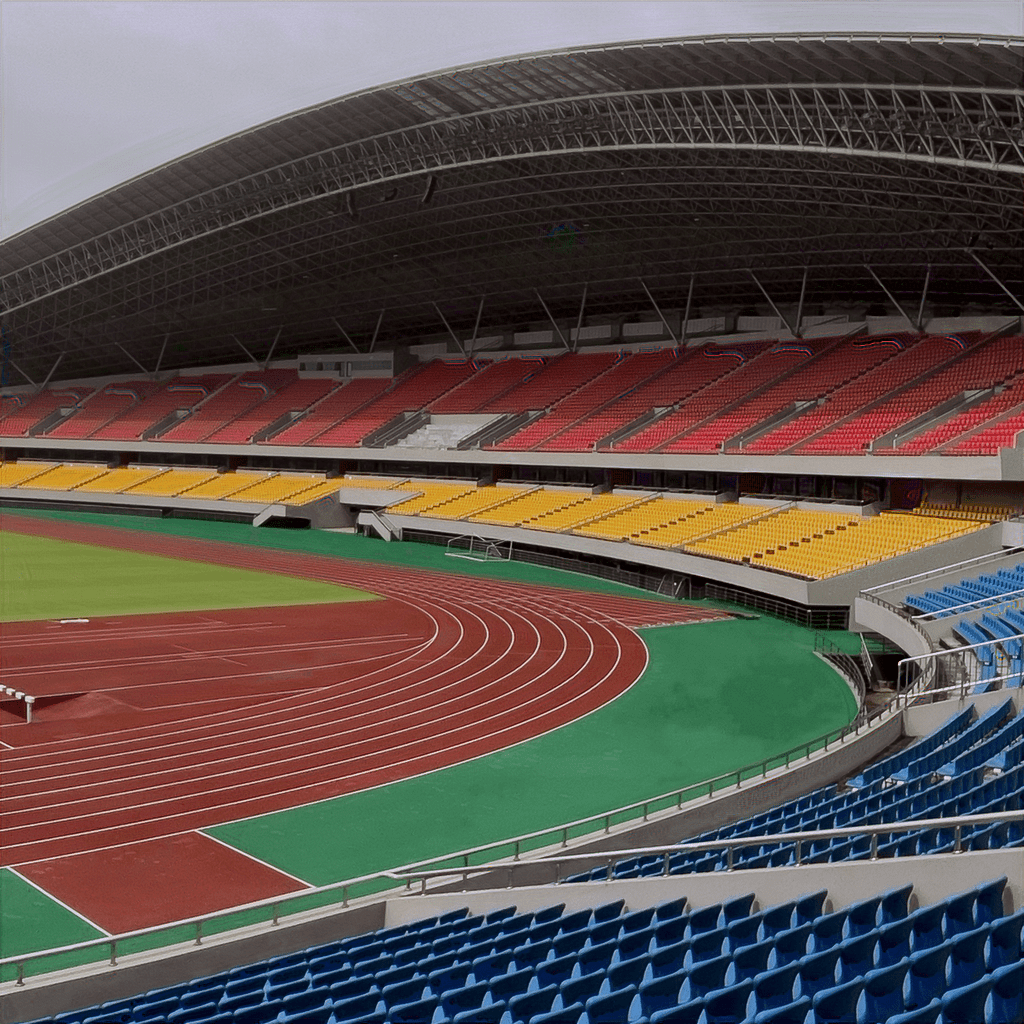}}
    \hspace{1pt}
    \subfloat[\footnotesize{CAN}]{\includegraphics[width=0.14\linewidth]{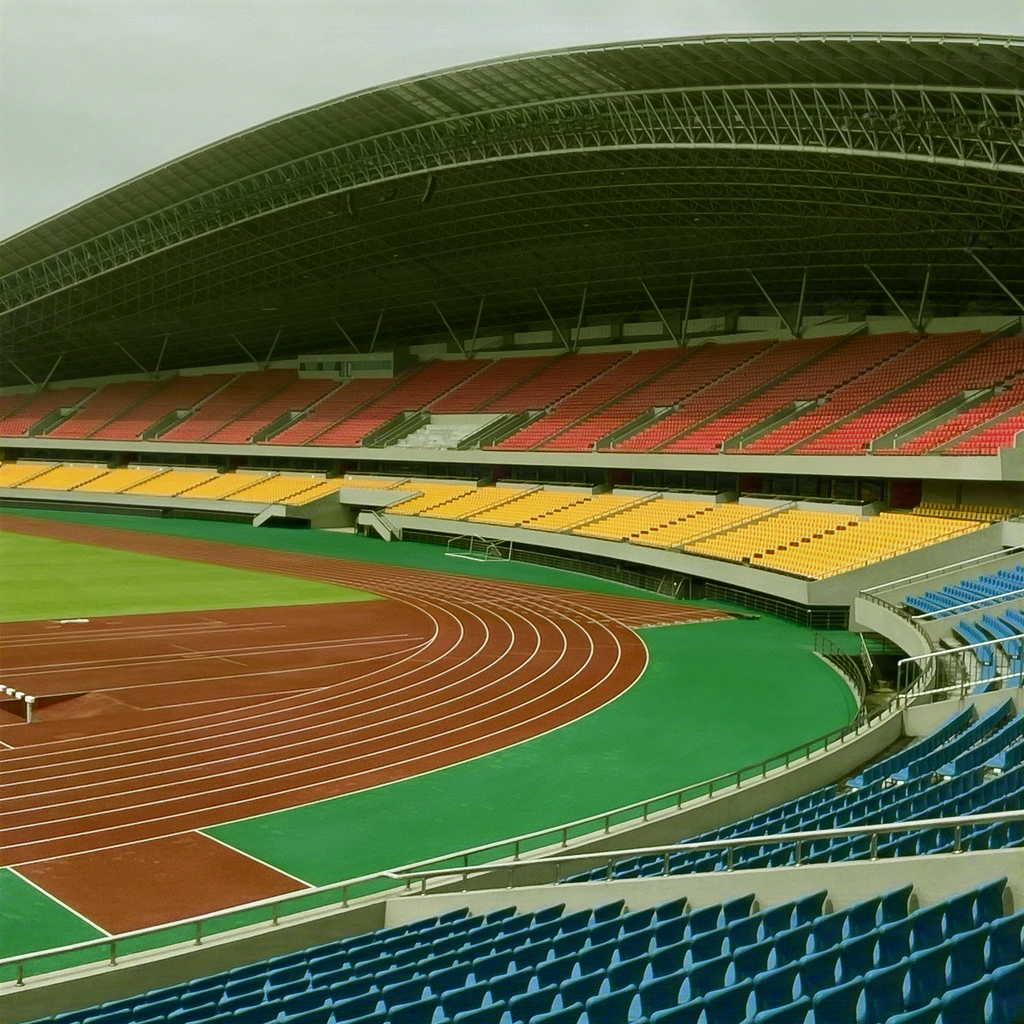}}
    \hspace{1pt}
    \subfloat[\footnotesize{CD-Net}]{\includegraphics[width=0.14\linewidth]{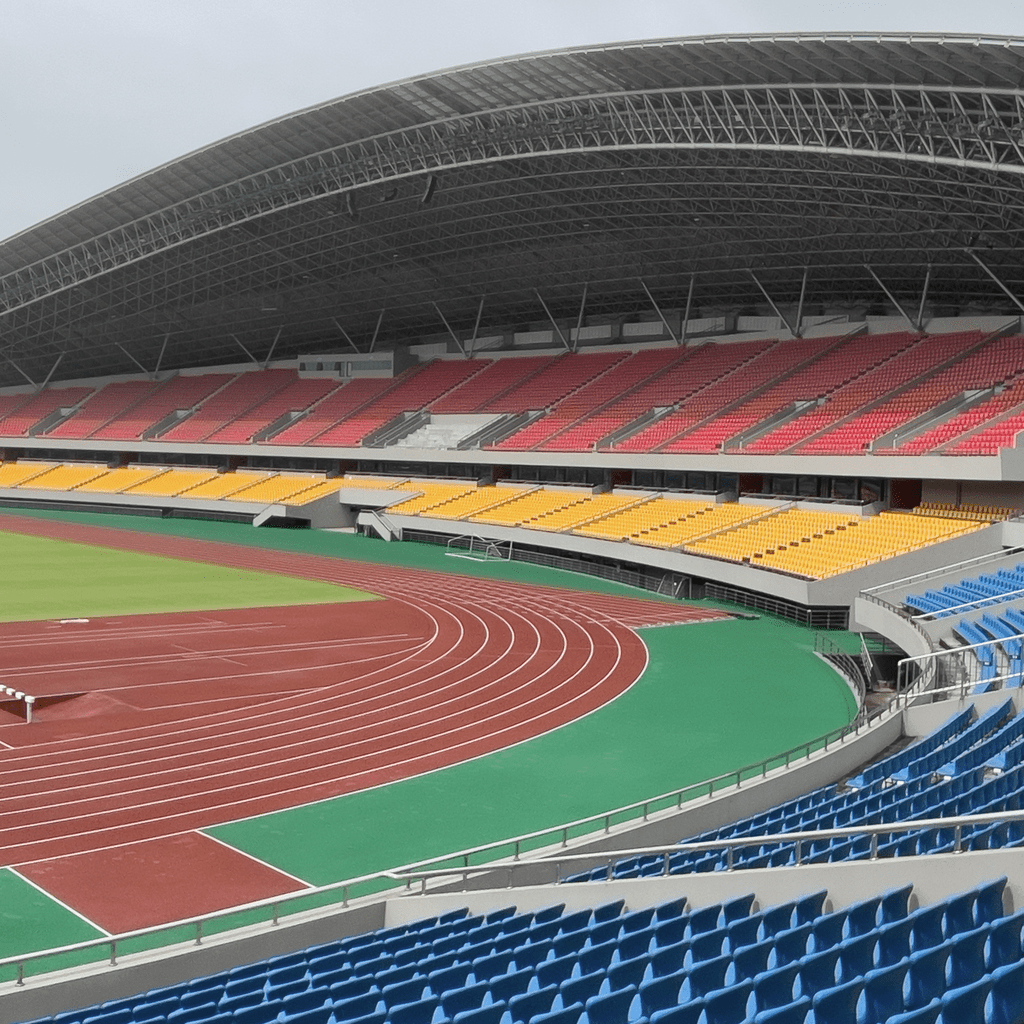}}
\end{minipage}
\caption{Reference image recovery. Starting from (b) a Gaussian noise image and (h) a tone-altered image of the reference, we recover images by optimizing the predicted CD to a reference image, using different DNN-based CD measures. (c)-(g) and (i)-(m) are recovered from the Gaussian noise image and the tone-altered image of the reference, respectively.}
\label{fig:image_recovery_test}
\end{figure*}

\subsection{Further Testing}\label{subsec:ft}
\noindent \textbf{Generalization to Unseen Alternations.} To test the generalizability of the proposed CD-Net to unseen alternations, we compare it against the competing methods on three color-related distortions (quantization noise, image color quantization with dither, and chromatic aberrations) sampled from an image quality dataset - TID2013~\cite{Ponomarenko2015image}. We report the comparison results in Table \ref{tab:comparision_tid2013}, from which we observed that CD-Net performs much better compared to pixel-wise CD metrics and their spatial extensions, but under-performs LPIPS and \FLIP, which may have been exposed to images of similar color appearances during training.

\noindent \textbf{Generalization to Homogeneous Color Patch Data.} We further test CD-Net on the COM dataset~\cite{luo2001ciede2000}, which is a combination of four color patch datasets, \ie, BFD-P~\cite{luo2001ciede2000}, Leeds~\cite{kim1997leeds}, Witt~\cite{witt1999witt}, and RIT-DuPont~\cite{berns1991rit-dupont}. For a fair comparison, we convert the color pairs in CIE XYZ values to $128\times128$ homogeneous color patches.

Table \ref{tab:color_patch_comparision} lists the comparison results of CD-Net against the competing methods. 
We find that CD-Net performs better than CIELAB on the COM dataset, despite not exposed to color patch data during training. This indicates that the high-dimensional ``color space'' learned directly from image pairs in SPCD seems to be more perceptually uniform than CIELAB, showing the promise of the proposed SPCD dataset in facilitating color research. As expected, our method underperforms CIEDE2000 on the COM dataset, since the latter is derived by deliberately fitting these datasets. 
Overall, CD-Net shows better generalizability compared to image-based CD methods, supporting its simple design philosophy.
\vspace{0.4em} \\
\noindent \textbf{Empirical Verification as a Proper Metric.} 
\label{sec:verification_of_mathematical_property}
As previously discussed, CD-Net is a proper metric in the transformed space, and it remains to be seen whether CD-Net is (or behaves like) a proper metric in RGB space. The non-negative and symmetric properties are immediately apparent from Eq. \eqref{eq:mds}. We empirically probe the identity of indiscernibles (\ie, $\varDelta E(x,y) = 0 \Longleftrightarrow x = y$) through the task of reference image recovery~\cite{ding2021comparison}. Given an RGB image $x$ and an initial image $y$, we try to recover $x$ by solving 
\begin{align}
    y^\star = \mathop{\arg\min}\limits_{y} \varDelta E(x,y),
    \label{eq:identity}
\end{align}    
where $\varDelta E(\cdot,\cdot)$ denotes a CD measure with a lower value indicating a smaller CD, and $y^\star$ is the recovered image. Fig.~\ref{fig:image_recovery_test} illustrates the recovery results starting from two different initializations - a Gaussian noise image and a tone-altered version of the reference image, respectively. For all CD measures, the optimization converges to the CD scores far below the JND (\ie, $\varDelta E_{ab}^{*} \approx 2.3$). We observe that CD-Net successfully recovers the reference image from both noise and tone-altered images. However, the four deep CNN-based methods generate final images completely different from the corresponding reference image, recovering limited structures and producing annoying distortions. Qualitatively, we find that our observation is consistent
for a wide range of natural photographic images with diverse content variability.
The results of reference image recovery provide additional strong evidence that deep CNN-based methods tend to overfit the training set, and may be less useful in perceptual optimization of computational methods for smartphone photography. 

We further empirically probe whether the triangle inequality (\ie, $\varDelta E(x,y) + \varDelta E(y,z) \ge \varDelta E(x,z) $) is satisfied by testing on nearly two million image triplets with the same content randomly generated from $15,335$ images described in Sec.~\ref{subsec:dcon}. We compare CD-Net with the four deep CNN-based methods, and find that only the VGG-based method violates the triangle inequality on thirteen triplets. No counterexamples are found for CD-Net and the remaining three CNN-based models. Putting  together, we empirically prove that CD-Net behaves as a proper metric.

\section{Conclusion}
In this paper, we have revisited the challenging and long-standing problem of CD assessment of natural photographic images, especially in the age of smartphone photography. From a data-driven perspective, we built a large-scale benchmark dataset consisting of a total of $30,000$ image pairs of human-rated CD scores, and trained a lightweight DNN, CD-Net, for reliable CD assessment. The proposed CD-Net has been demonstrated to predict the CDs of natural images accurately, offer competitive CD maps for potential use in local color manipulation, generalize reasonably to homogeneous color patch data, and behave as a proper metric in the mathematical sense.   
We hope that our newly established dataset can become valuable resource for further developing CD metrics, and that our CD-Net can benefit related fields in smartphone photography.  

{\small
\bibliographystyle{IEEEtran}
\bibliography{egbib}
}
\end{document}